\documentclass[journal]{IEEEtran}

\usepackage[utf8]{inputenc}
\usepackage{color}
\usepackage{xcolor}
\usepackage{array}
\usepackage{verbatim}
\usepackage{float}
\usepackage{amsmath}
\usepackage{amsthm}
\usepackage{amssymb}
\usepackage{graphicx}
\usepackage{longtable}
\usepackage{multirow}
\usepackage{booktabs}
\usepackage[unicode=true,
bookmarks=false,
breaklinks=false,pdfborder={0 0 1},colorlinks=false]
{hyperref}
\hypersetup{
	colorlinks,bookmarksopen,bookmarksnumbered,citecolor=blue,urlcolor=blue}
\usepackage{cite}

\usepackage{lipsum}
\usepackage{mathtools}
\usepackage{cuted}
\providecommand{\tabularnewline}{\\}
\usepackage{algorithmic}
\usepackage{longtable}

\floatstyle{ruled}
\newfloat{algorithm}{tbp}{loa}
\providecommand{\algorithmname}{Algorithm}
\floatname{algorithm}{\protect\algorithmname}

\makeatletter
\let\oldforeign@language\foreign@language
\DeclareRobustCommand{\foreign@language}[1]{%
	\lowercase{\oldforeign@language{#1}}}

\let\oldforeign@language\foreign@language
\DeclareRobustCommand{\foreign@language}[1]{%
	\lowercase{\oldforeign@language{#1}}}

\ifCLASSINFOpdf
\else
\fi

\@ifundefined{showcaptionsetup}{}{%
	\PassOptionsToPackage{caption=false}{subfig}}
\usepackage{subfig}

\usepackage{balance}

\ifCLASSINFOpdf
\else
\fi

\newtheorem{rem}{Remark}

\ifCLASSINFOpdf
\else
\fi

\def\ps@IEEEtitlepagestyle{%
	\def\@oddhead{\parbox[t][\height][t]{\textwidth}{\centering \scriptsize
			Personal use of this material is permitted. Permission from the author(s) and/or copyright holder(s), must be obtained for all other uses. Please contact us and provide details if you believe this document breaches copyrights.\\
			\noindent\makebox[\linewidth]{}
		}\hfil\hbox{}}%
	\def\@evenhead{\scriptsize\thepage \hfil \leftmark\mbox{}}%
	\def\@oddfoot{\parbox[t][\height][l]{\textwidth}{
			\vspace{-20pt}{\rule{\textwidth}{0.4pt}}\\ \footnotesize{\bf{\footnotesize\textcolor{red}{K. Ghanizadegan and H. A. Hashim, "Quaternion-based Unscented Kalman Filter for 6-DoF Vision-based Inertial Navigation in GPS-denied Regions," IEEE Transactions on Instrumentation and Measurement, 2025.}}} doi: \href{https://doi.org/10.1109/TIM.2024.3509582}{10.1109/TIM.2024.3509582}\\
			\noindent\makebox[\linewidth]
		}\hfil\hbox{}}%
	\def\@evenfoot{\MYfooter}}

\makeatother
\begin{document}
	\bstctlcite{IEEEexample:BSTcontrol}

\title{Quaternion-based Unscented Kalman Filter for 6-DoF Vision-based Inertial Navigation in GPS-denied Regions}

\author{Khashayar Ghanizadegan and Hashim A. Hashim
\thanks{This work was supported in part by National Sciences and Engineering Research Council of Canada (NSERC), under the grants RGPIN-2022-04937 and DGECR-2022-00103.} 
	\thanks{K. Ghanizadegan and H. A. Hashim are with the Department of Mechanical
		and Aerospace Engineering, Carleton University, Ottawa, Ontario, K1S-5B6,
		Canada (e-mail: hhashim@carleton.ca).}
}



\maketitle
\begin{abstract}
This paper investigates the orientation, position, and linear velocity
estimation problem of a rigid-body moving in three-dimensional (3D)
space with six degrees-of-freedom (6 DoF). The highly nonlinear navigation
kinematics are formulated to ensure global representation of the navigation
problem. A computationally efficient Quaternion-based Navigation Unscented
Kalman Filter (QNUKF) is proposed on $\mathbb{S}^{3}\times\mathbb{R}^{3}\times\mathbb{R}^{3}$
imitating the true nonlinear navigation kinematics and utilize onboard
Visual-Inertial Navigation (VIN) units to achieve successful GPS-denied
navigation. The proposed QNUKF is designed in discrete form to operate
based on the data fusion of photographs garnered by a vision unit
(stereo or monocular camera) and information collected by a low-cost
inertial measurement unit (IMU). The photographs are processed to
extract feature points in 3D space, while the 6-axis IMU supplies
angular velocity and accelerometer measurements expressed with respect
to the body-frame. Robustness and effectiveness of the proposed QNUKF
have been confirmed through experiments on a real-world dataset collected
by a drone navigating in 3D and consisting of stereo images and 6-axis
IMU measurements. Also, the proposed approach is validated against
standard state-of-the-art filtering techniques.
\end{abstract}

\begin{IEEEkeywords}
Localization, Navigation, Unmanned Aerial Vehicle, Sensor-fusion,
Inertial Measurement Unit, Vision Unit.
\end{IEEEkeywords}

\rule{0.47\textwidth}{1pt}\\
For video of navigation experiment visit: \href{https://youtu.be/CP3xiOcGrTc}{link}\vspace{4pt}
\\
For video of navigation experiment comparison visit: \href{https://youtu.be/BWraOI0LAXo}{link}
\\
\vspace{-1pt}
\rule{0.49\textwidth}{1pt}

\section{Introduction}\label{sec1}

\subsection{Motivation}
\IEEEPARstart{N}{avigation} is generally defined as the process of determining the
position, orientation, and linear velocity of a rigid-body in space,
performed by either an observer or external entities. Navigation,
both in its complete and partial forms, plays a crucial role in numerous
systems and fields, significantly improving operational performance
and efficiency. Its applications range from robotics to smartphones,
aerospace, and marines \cite{hashim2021gps,Xu2023ExtendedNoise},
highlighting its vital importance in driving technological and scientific
progress. In the smartphone industry, the precise estimation of a
pedestrian's position and heading through their mobile device is essential
for improving location-based services and facilitating seamless navigation
between outdoor and indoor environments. Advancing accuracy and reliability
of the estimation is critical for enriching user experiences in wayfinding
applications, as documented in recent studies \cite{hashim2023observer,wang2020pedestrian,hashim2021geometricNAV,Asraf2022PDRNet:Framework,Jiang2022ImplementationSmartphone}.
In aerospace, the precise estimation of satellite orientation is pivotal
for the functional integrity of satellites, particularly for the accurate
processing of observational data. This aspect is fundamental to ensuring
the reliability and effectiveness of satellite operations \cite{hashim2019Ito,aerospace7010003,hashim2020systematic}.

Autonomous and unmanned robots employ navigation techniques to enhance
the precision of Global Navigation Satellite System (GNSS)-based localization
systems, such as the Global Positioning System (GPS) and GLONASS.
These enhancements are critical for ensuring operational functionality
in environments where GPS signals are obstructed or unavailable \cite{hashim2021geometricNAV,Scaramuzza2014VisionControlled}.
This capability is particularly vital for Unmanned Aerial Vehicles
(UAVs) navigating indoors or within densely constructed urban areas,
where direct line-of-sight to satellites is frequently obscured. Navigation
methods that do not rely on GNSS hold significant importance in maritime
contexts, where GNSS signals can be unreliable or entirely absent,
especially in deep waters or in the vicinity of harbors \cite{Tan2011A}.
Such GNSS-independent navigation techniques are crucial for ensuring
the safety and efficiency of marine operations, where traditional
satellite-based positioning systems may not provide the necessary
accuracy or reliability.

\subsection{Related Work}

A naive approach to localizing an agent in a GNSS-denied environment
is to apply Dead Reckoning (DR) using a 6-axis Inertial Measurement
Unit (IMU) rigidly attached to the vehicle and comprised of an accelerometer
(supplying the apparent acceleration measurements) and a gyroscope
(supplying angular velocity measurements) \cite{hashim2021geometricNAV,hashim2021gps}.
DR is able to produce the vehicle navigation state utilizing gyroscope
measurements and integration of acceleration provided that initial
navigation state is available \cite{RostonDeadRobots,hashim2021geometricNAV}.
DR is widely utilized for pedestrian position and heading estimation
in scenarios where GNSS is unreliable (e.g., indoor environments).
Typically, the step count, step length, and heading angle are each
determined by an independent sensor, and these measurements are then
incorporated into the last known position to estimate the new position.
This method serves as a preprocessing step for the DR, and aims to
enhance the accuracy of heading estimation. However, a major limitation
of DR is the accumulation of errors in the absence of any measurement
model, leading to drift over time. In the pedestrian DR domain, multiple
solutions have been proposed to mitigate the drift issue, such as
collection of distance information from known landmarks, a deep-learning
based approach to supply filtered acceleration data \cite{Asraf2022PDRNet:Framework},
Recurrent Neural Network (RNN) with adaptive estimation of noise covariance
\cite{Zhou2022IMUAlgorithm} and others. These approaches are subsequently
coupled with a Kalman-type filter to propagate the state estimate.
However, these methods, while enhancing the robustness of the filter,
fail to correct the state estimate if significant drift occurs.

In scenarios where a robot navigates in a known environment, leveraging
measurements from the environment can be a solution to the drift problem
encountered by the pure IMU-based navigation algorithms. Ultra-wideband
(UWB) has recently gained popularity as an additional sensor in the
navigation suite. A robot equipped with a UWB tag gains knowledge
of its positions by communicating with fixed UWB anchors with known
positions \cite{Hashim2023NonlinearFusion,Xu2023ExtendedNoise,Mi2023ConstrainedNavigation,helgesen2022inertial,You2020DataUAV}.
This is beneficial for structured environments with sufficient fixed
reference points (e.g., ship docking \cite{helgesen2022inertial}
and warehouse \cite{Hashim2023NonlinearFusion}). However, the need
for infrastructure with known fixed anchors limits the generalizability
of UWB-based navigation algorithms in unstructured regions \cite{hashim2023observer}.
With advances in three-dimensional (3D) point cloud registration and
processing technologies, such as the Iterative Closest Point (ICP)
\cite{121791} and Coherent Point Drift (CPD) \cite{Myronenko2009PointSR},
the dependency on initial environment knowledge is significantly reduced.
Utilizing these algorithms, a robot equipped only with its perspective
of 3D points at multiple steps can reconstruct the environment. This
reconstructed environment then is treated as known, allowing the feature
points obtained from the environment to serve as independent measurements
alongside the IMU \cite{hashim2021gps,hashim2021geometric}. Technologies
such as Sound Navigation And Ranging (SONAR), Light Detection And
Ranging (LIDAR), and visual inputs from mono or stereo cameras exemplify
such measurements. SONAR and LIDAR sensors emit mechanical and electromagnetic
waves, respectively, and construct 3D point clouds by measuring the
distances that each wave travels. SONAR is a popular sensor in marine
applications \cite{bai2024side}, as water propagates mechanical waves
(sound waves) significantly better than electromagnetic waves (light
waves). LIDAR is commonly used in space application \cite{christian2013survey}
since sound waves cannot propagate through space while light can.
However, both LIDAR and SONAR struggle in complex indoor and urban
environments due to their inability to capture colour and texture.
Therefore, vision-based aided navigation (stereo or monocular camera
+ IMU) has emerged as a reliable navigation alternative for GPS-denied
regions. In particular, modern UAVs are equipped with high-resolution
cameras which are more cost-effective and computationally reliable.

\subsection{Vision-based Navigation and Contributions}

Feature coordinates in 3D space can be extracted using a monocular
camera (given a sequence of two images with persistent vehicle movement)
or a stereo (binocular) camera \cite{hashim2021geometricNAV,wei2022optimization}.
While stereo vision-based navigation algorithms, such as those proposed
in \cite{murray2000using} and \cite{huntsberger2011stereo}, generally
outperform monocular-based systems, they encounter difficulties in
dark environments and in absence of clear features, such as a drone
facing a blank wall. Fusing data from IMUs and vision sensors is a
widely accepted solution to address navigation challenges, as these
modalities complement each other \cite{hashim2021geometricNAV,kim2007real,huang2022vwr,mourikis2007multi}.
This integration of vision sensors and IMU is commonly referred to
as Visual-Inertial Navigation (VIN) \cite{huang2019visual}. Kalman-type
filters have been extensively used to address the VIN problem due
to their low computational requirements and ability to handle non-linear
models via linearization. An early significant contribution in this
area was made by Mourikis et al. \cite{mourikis2007multi}, who developed
a multi-state Kalman filter-based algorithm for a VIN system that
operates based on the error state dynamics. This approach allows the
orientation, which has three degrees of freedom, to be represented
by a three-dimensional vector. To address the issue of computational
complexity and expand the algorithm proposed in \cite{mourikis2007multi},
the work in \cite{Erdem2015FusingTracking} explored various methods
for fusing IMU and vision data using the Extended Kalman Filter (EKF).
Specifically, they examined the integration of these data sources
during the EKF prediction and update phases. Although EKF is an efficient
algorithm, its major shortcoming is the use of local linearization
which disregards the high nonlinearity of the navigation kinematics.
Therefore, there is a need for multiple interacting models to tackle
EKF shortcomings \cite{khalkhali2019multi}. Moreover, EKF is designed
to account only for white noise attached to the measurements, making
it unfit for measurements with colored noise \cite{hashim2019Ito}.

\paragraph*{Contributions}Unscented Kalman Filter (UKF) generally
outperforms the EKF \cite{hashim2019Ito} since it can accurately
capture the nonlinear kinematics while handling white and colored
noise \cite{hashim2019Ito,ref:ukf}. Motivated by the above discussion,
the contributions of this work are as follows: (1) Novel geometric
Quaternion-based Navigation Unscented Kalman Filter (QNUKF) is proposed
on $\mathbb{S}^{3}\times\mathbb{R}^{3}\times\mathbb{R}^{3}$ in discrete
form mimicking the true nonlinear navigation kinematics of a rigid-body
moving in the 6 DoF; (2) The proposed QNUKF is tailored to supply
attitude, position, and linear velocity estimates of a rigid-body
equipped with a VIN unit and applicable for GPS-denied navigation;
(3) The computational efficiency and robustness of the proposed QNUKF
is tested using a real-world dataset collected by a quadrotor flying
in the 6 DoF at a low sampling rate and subsequently compared to state-of-the-art
industry standard filtering techniques.

\subsection{Structure}

The remainder of the paper is organized as follows: Section \ref{sec:Preliminaries-and-Math}
discusses preliminary concepts and mathematical foundations; Section
\ref{sec:Problem-Formulation} formulates the nonlinear navigation
kinematics problem; Section \ref{sec:Filter} introduces the structure
of the novel QNUKF; Section \ref{sec:Results} validates the algorithm's
performance using a real-world dataset; and Section \ref{sec:Conclusion}
provides concluding remarks.

\section{Preliminaries\label{sec:Preliminaries-and-Math}}

\begin{table}[t]
	\centering{}\caption{\label{tab:Table-of-Notations2}Nomenclature}
	\begin{tabular}{>{\raggedright}p{2cm}l>{\raggedright}p{5.6cm}}
		\toprule 
		\addlinespace[0.1cm]
		$\left\{ \mathcal{B}\right\} $ / $\left\{ \mathcal{W}\right\} $ & : & Fixed body-frame / fixed world-frame\tabularnewline
		\addlinespace[0.1cm]
		$\mathbb{SO}\left(3\right)$ & : & Special Orthogonal Group of order 3\tabularnewline
		\addlinespace[0.1cm]
		$\mathbb{S}^{3}$ & : & Three-unit-sphere\tabularnewline
		\addlinespace[0.1cm]
		$\mathbb{Z}$, $\mathbb{Z}^{+}$ & : & Integer and positive integer space\tabularnewline
		\addlinespace[0.1cm]
		$q_{k},\hat{q}_{k}$ & : & True and estimated quaternion at step $k$\tabularnewline
		\addlinespace[0.1cm]
		$p_{k},\hat{p}_{k}$ & : & True and estimated position at step $k$\tabularnewline
		\addlinespace[0.1cm]
		$v_{k},\hat{v}_{k}$ & : & True and estimated linear velocity at step $k$\tabularnewline
		\addlinespace[0.1cm]
		$r_{e,k}$, $p_{e,k}$, $v_{e,k}$ & : & Attitude, position, and velocity estimation error\tabularnewline
		\addlinespace[0.1cm]
		$a_{k},a_{m,k}$ & : & True and measured acceleration at step $k$\tabularnewline
		\addlinespace[0.1cm]
		$\omega_{k},\omega_{m,k}$ & : & True and measured angular velocity at step $k$\tabularnewline
		\addlinespace[0.1cm]
		$n_{\omega,k},n_{a,k}$ & : & Angular velocity and acceleration measurements noise\tabularnewline
		\addlinespace[0.1cm]
		$b_{\omega,k},b_{a,k}$ & : & Angular velocity and acceleration measurements bias\tabularnewline
		\addlinespace[0.1cm]
		$C_{\times}$ & : & Covariance matrix of $n_{\times}$.\tabularnewline
		\addlinespace[0.1cm]
		$f_{b},f_{w}$ & : & Feature points coordinates in $\left\{ \mathcal{B}\right\} $ and
		$\left\{ \mathcal{W}\right\} $.\tabularnewline
		\addlinespace[0.1cm]
		$x_{k}$, $x_{k}^{a}$, $u_{k}$ & : & The state, augmented state, and input vectors at the $k$th time step\tabularnewline
		\addlinespace[0.1cm]
		$\hat{z}_{k},z_{k}$ & : & Predicted and true measurement\tabularnewline
		\addlinespace[0.1cm]
		$\{\mathcal{X}_{i|j}\}$, $\{\mathcal{X}_{i|j}^{a}\}$, $\{\mathcal{Z}_{i|j}\}$ & : & Sigma points of state, augmented state, and measurements\tabularnewline
		\bottomrule
	\end{tabular}
\end{table}

\paragraph*{Notation}In this paper, the set of integers, positive
integers, and $m_{1}$-by-$m_{2}$ matrices of real numbers are represented
by $\mathbb{Z}$, $\mathbb{Z}^{+}$, and $\mathbb{R}^{m_{1}\times m_{2}}$,
respectively. The Euclidean norm of a vector $v\in\mathbb{R}^{m_{v}}$
is defined by $||v||=\sqrt{v^{\top}v}$, where $v^{\top}$ refers
to the transpose of $v$. $\mathbf{I}_{m}$ is identity matrix with
dimension $m$-by-$m$. The world-frame $\left\{ \mathcal{W}\right\} $
refers to a reference-frame fixed to the Earth and $\left\{ \mathcal{B}\right\} $
describes the body-frame which is fixed to the vehicle's body. Table
\ref{tab:Table-of-Notations2} provides important notation used throughout
the paper. The skew-symmetric $[\cdot]_{\times}$ of $v\in\mathbb{R}^{3}$
is defined by:
\begin{equation}
	[v]_{\times}=\left[\begin{array}{ccc}
		0 & -v_{3} & v_{2}\\
		v_{3} & 0 & -v_{1}\\
		-v_{2} & v_{1} & 0
	\end{array}\right]\in\mathfrak{so}(3),\hspace{1em}v=\left[\begin{array}{c}
		v_{1}\\
		v_{2}\\
		v_{3}
	\end{array}\right]
\end{equation}
The operator ${\rm vex}(\cdot)$ is the inverse mapping skew-symmetric
operator to vector ${\rm vex}:\mathfrak{so}(3)\rightarrow\mathbb{R}^{3}$
\begin{equation}
	{\rm vex}([v]_{\times})=v\in\mathbb{R}^{3}\label{eq:vex}
\end{equation}
The anti-symmetric projection operator $\mathcal{P}_{a}(\cdot):\mathbb{R}^{3\times3}\rightarrow\mathfrak{so}(3)$
is defined as:
\begin{equation}
	\mathcal{P}_{a}(M)=\frac{1}{2}(M-M^{\top})\in\mathfrak{so}(3),\hspace{0.5em}\forall M\in\mathbb{R}^{m\times m}\label{eq:pa}
\end{equation}

\subsection{Orientation Representation}

$\mathbb{SO}(3)$ denotes the Special Orthogonal Group of order 3
and is defined by: 
\begin{equation}
	\mathbb{SO}(3):=\left\{ \left.R\in\mathbb{R}^{3\times3}\right|det(R)=+1,RR^{\top}=\mathbf{I}_{3}\right\} 
\end{equation}
where $R\in\mathbb{SO}(3)$ denotes orientation of a rigid-body. Euler
angle parameterization provides an intuitive representation of the
rigid-body's orientation in 3D space, often considered analogous to
roll, pitch, and yaw angles around the $x$, $y$, and $z$ axes,
respectively. This representation has been adopted in many pose estimation
and navigation problems such as \cite{jiang2022iterative,bai2024side}.
However, this representation is kinematically singular and not globally
defined. Thus, Euler angles fail to represent the vehicle orientation
in several configurations and are subject to singularities \cite{hashim2019special}.
Angle-axis and Rodrigues vector parameterization are also subject
to singularity in multiple configurations \cite{hashim2019special}.
Unit-quaternion provides singularity-free orientation representation,
while being intuitive and having only 4 parameters with 1 constraint
to satisfy the 3 DoF \cite{hashim2019special}. A quaternion vector
$q$ is defined in the scaler-first format by $q=[q_{w},q_{x},q_{y},q_{z}]^{\top}=[q_{w},q_{v}^{\top}]^{\top}\in\mathbb{S}^{3}$
with $q_{v}\in\mathbb{R}^{3}$, $q_{w}\in\mathbb{R}$, and
\begin{equation}
	\mathbb{S}^{3}:=\left\{ q\in\left.\mathbb{R}^{4}\right|||q||=1\right\} 
\end{equation}
Let $\otimes$ denote quaternion product of two quaternion vectors.
The quaternion product of two rotations $q_{1}=[q_{w1},q_{v1}]^{\top}$
and $q_{2}=[q_{w2},q_{v2}]^{\top}$ is given by \cite{hashim2019special}:
\begin{align}
	q_{3} & =q_{1}\otimes q_{2}\nonumber \\
	& =\begin{bmatrix}q_{w1}q_{w2}-q_{v1}^{\top}q_{v2}\\
		q_{w1}q_{v2}+q_{w2}q_{v1}+[q_{v1}]_{\times}q_{v2}
	\end{bmatrix}\in\mathbb{S}^{3}\label{eq:qxq}
\end{align}
The inverse quaternion of $q=[q_{w},q_{x},q_{y},q_{z}]^{\top}=[q_{w},q_{v}^{\top}]^{\top}\in\mathbb{S}^{3}$
is represented by
\begin{equation}
	q^{-1}=[q_{w},-q_{x},-q_{y},-q_{z}]^{\top}=[q_{w},-q_{v}^{\top}]^{\top}\in\mathbb{S}^{3}\label{eq:q_inv}
\end{equation}
Note that $q_{I}=[1,0,0,0]^{\top}$ refers to the quaternion identity
where $q\otimes q^{-1}=q_{I}$. The rotation matrix can be described
using quaternion $R_{q}(q)$ such that $R_{q}:\mathbb{S}^{3}\rightarrow\mathbb{SO}(3)$
\cite{hashim2019special}:
\begin{equation}
	R_{q}(q)=(q_{w}^{2}-||q_{v}||^{2})I_{3}+2q_{v}q_{v}^{\top}+2q_{w}[q_{v}]_{\times}\in\mathbb{SO}(3)\label{Q2R}
\end{equation}
Also, quaternion can be extracted given a rotation matrix $R\in\mathbb{SO}(3)$
and the mapping $q_{r}:\mathbb{SO}(3)\rightarrow\mathbb{S}^{3}$ is
defined by \cite{hashim2019special}:
\begin{equation}
	q_{r}(R)=\left[\begin{array}{c}
		q_{w}\\
		q_{x}\\
		q_{y}\\
		q_{z}
	\end{array}\right]=\left[\begin{array}{c}
		\frac{1}{2}\sqrt{1+R_{(1,1)}+R_{(2,2)}+R_{(3,3)}}\\
		\frac{1}{4q_{w}}(R_{(3,2)}-R_{(2,3)})\\
		\frac{1}{4q_{w}}(R_{(1,3)}-R_{(3,1)})\\
		\frac{1}{4q_{w}}(R_{(2,1)}-R_{(1,2)})
	\end{array}\right]\label{R2Q}
\end{equation}
The rigid-body's orientation can be extracted using angle-axis parameterization
through a unity vector $u=[u_{1},u_{2},u_{3}]\in\mathbb{S}^{2}$ rotating
by an angle $\theta\in\mathbb{R}$ \cite{hashim2019special} where
$\mathbb{S}^{2}:=\{\left.u\in\mathbb{R}^{3}\right|||u||=1\}$. The
rotation vector $r$ can be defined using the angle-axis parameterization
such that $r_{\theta,u}:\mathbb{R}\times\mathbb{S}^{2}\rightarrow\mathbb{R}^{3}$:
\begin{equation}
	r=r_{\theta,u}(\theta,u)=\theta u\in\mathbb{R}^{3}\label{AA2r}
\end{equation}
The rotation matrix can be extracted given the rotation vector and
the related map is given by  $R_{r}:\mathbb{R}^{3}\rightarrow\mathbb{SO}(3)$
\cite{hashim2019special}
\begin{align}
	R_{r}(r) & =\exp([r]_{\times})\in\mathbb{SO}(3)\nonumber \\
	& =\mathbf{I}_{3}+\sin(\theta)\left[u\right]_{\times}+\left(1-\cos(\theta)\right)\left[u\right]_{\times}^{2}\label{r2R}
\end{align}
The unity vector and rotation angle can be obtained from rotation
matrix $\theta_{R},u_{R}:\mathbb{SO}(3)\rightarrow\mathbb{R}^{3}\times\mathbb{R}$
\cite{hashim2019special}:
\begin{equation}
	\left\{ \begin{aligned}\theta_{R} & =\arccos\left(\frac{{\rm Tr}(R)-1}{2}\right)\in\mathbb{R}\\
		u_{R} & =\frac{1}{\sin\theta_{R}}{\rm vex}(\mathcal{P}_{a}(R))\in\mathbb{S}^{2}
	\end{aligned}
	\right.\label{R2AA}
\end{equation}
Given the rotation vector $r$ in \eqref{AA2r}, in view of \eqref{r2R}
and \eqref{R2Q}, one has \cite{hashim2019special}
\begin{align}
	q_{r}(r) & =q_{R}\left(R_{r}(r)\right)\in\mathbb{S}^{3}\nonumber \\
	& =\left[\cos(\theta/2),\sin(\theta/2)u^{\top}\right]^{\top}\in\mathbb{S}^{3}\label{eq:r2q}
\end{align}
Finally, to find the rotation vector $r_{q}(q)$ corresponding to
the quaternion $q$, equations \eqref{Q2R}, \eqref{R2AA}, and \eqref{AA2r}
are used as follows:
\begin{equation}
	r_{q}(q)=r_{\theta,u}(\theta_{R}(R_{q}(q)),u_{R}(R_{q}(q)))\in\mathbb{R}^{3}\label{eq:q2r}
\end{equation}

\subsection{Summation, Deduction Operators, and Weighted Average}

Quaternions and rotation vectors cannot be directly added or subtracted
in a straightforward manner, whether separately or side-dependent.
Let us define the side-dependent summation $\oplus$ and subtraction
$\ominus$ operators to enable quaternion and rotation vector summation.
In view of \eqref{eq:vex}, \eqref{eq:q_inv}, \eqref{AA2r}, and
\eqref{eq:r2q}, one has:
\begin{align}
	q\oplus r & :=q_{r}(r)\otimes q\in\mathbb{S}^{3}\label{eq:q+r}\\
	q\ominus r & :=q_{r}(r)^{-1}\otimes q\in\mathbb{S}^{3}\label{eq:q-r}
\end{align}
where $q\in\mathbb{S}^{3}$ denotes quaternion vector and $r\in\mathbb{R}^{3}$
denotes rotation vector. The equations \eqref{eq:q+r} and \eqref{eq:q-r}
provide the resultant quaternion after subsequent rotations represented
by $q$ combined with $r$, and $q$ combined with the inverse of
$r$, respectively.  Using \eqref{eq:qxq}, \eqref{eq:q_inv}, and
\eqref{eq:q2r}, the subtraction operator of $q_{1}\in\mathbb{S}^{3}$
and $q_{2}\in\mathbb{S}^{3}$ maps $\mathbb{S}^{3}\times\mathbb{S}^{3}\rightarrow\mathbb{R}^{3}$
and is defined as follows:
\begin{equation}
	q_{1}\ominus q_{2}:=r_{q}(q_{1}\otimes q_{2}^{-1})\in\mathbb{R}^{3}\label{eq:q-q}
\end{equation}
The expression \eqref{eq:q-q} provides the rotation vector that represents
the orientation error between the two quaternions $q_{1}$ and $q_{2}$.
Note that the expressions in \eqref{eq:q+r}, \eqref{eq:q-r}, and
\eqref{eq:q-q} do not only resolve the addition and subtraction operations
in the $\mathbb{S}^{3}$ and $\mathbb{R}^{3}$ spaces, but also provide
physical meaning for these operations, specifically in terms of subsequent
orientations. The expression \eqref{eq:q-q} provides the rotation
vector that represents the orientation error between the two quaternions
$q_{1}$ and $q_{2}$. The algorithms provided in \cite{hashim2019Ito,markley2007quaternion}
are used to obtain weighted mean ${\rm WM}(Q,W)$ of a set of quaternions
$Q=\{q_{i}\in\mathbb{S}^{3}\}$ and scaler weights $W=\{w_{i}\}$.
In order to calculate this weighted average, first the $4$-by-$4$
matrix $M$ is found by:
\[
M=\sum w_{i}q_{i}q_{i}^{\top}\in\mathbb{R}^{4\times4}
\]
Next, the unit eigenvector corresponding to the eigenvalue with the
highest magnitude is regarded as the weighted average. In other words:
\begin{equation}
	{\rm WM}(Q,W)=\text{EigVector}(M)_{i}\in\mathbb{S}^{3}\label{eq:weighted average}
\end{equation}
where
\[
i=\text{argmax}(|\text{EigValue}(M)_{i}|)\in\mathbb{R}
\]
with $\text{EigVector}(M)_{i}$ in \eqref{eq:weighted average} and
$\text{EigValue}(M)_{i}$ being the $i$th eigenvector and eigenvalue
of $M$, respectively.

\subsection{Probability}

The probability of a Random Variable (RV) $A$ being equal to $a$,
where $a\in\mathbb{R}^{m_{A}}$, is denoted by $\mathbb{P}(A=a)$,
or more compactly, $\mathbb{P}(a)$. Consider another RV $B$ with
a possible value $b\in\mathbb{R}^{m_{B}}$. The conditional probability
of $A=a$ given that $B=b$, denoted by $\mathbb{P}(a|b)$, can be
expressed as:
\[
\mathbb{P}(a|b)=\frac{\mathbb{P}(a,b)}{\mathbb{P}(b)}
\]
An $m$-dimensional RV $V\in\mathbb{R}^{m}$ drawn from a Gaussian
distribution with a mean of $\overline{V}\in\mathbb{R}^{m}$ and a
covariance matrix of $P_{v}\in\mathbb{R}^{m\times m}$ is represented
by the following:
\[
V\sim\mathcal{N}(\overline{V},P_{v})
\]
The Gaussian probability density function of $V$ is formulated below:
\begin{align*}
	\mathbb{P}(V) & =\mathcal{N}(V|\overline{V},P_{v})\\
	& =\frac{\exp\left(-\frac{1}{2}(V-\overline{V})^{\top}P_{v}^{-1}(V-\overline{V})\right)}{\sqrt{(2\pi)^{m}\det(P_{v})}}\in\mathbb{R}
\end{align*}
Let:
\begin{equation}
	\begin{pmatrix}A\\
		B
	\end{pmatrix}\sim\mathcal{N}\left(\begin{pmatrix}\overline{A}\\
		\overline{B}
	\end{pmatrix},\begin{pmatrix}P_{A} & P_{A,B}\\
		P_{A,B}^{\top} & P_{B}
	\end{pmatrix}\right)\label{eq:AB-normal}
\end{equation}
Then, given \eqref{eq:AB-normal}, the conditional probability function
$\mathbb{P}(A|B)$ can be calculated as:
\begin{align}
	\mathbb{P}(A|B) & =\mathcal{N}(A,B|\overline{A}+P_{A,B}P_{B}^{-1}(B-\overline{B}),\nonumber \\
	& \hspace{9em}P_{A}-P_{A,B}P_{B}^{-1}P_{A,B}^{\top})\label{eq:AB}
\end{align}

\subsection{The Unscented Transform}

Consider the square symmetric semi-positive definite matrix $M\in\mathbb{R}^{m_{M}\times m_{M}}$.
Using Singular Value Decomposition (SVD), let $U$, $S$, and $V\in\mathbb{R}^{m_{M}\times m_{M}}$
be the matrices of left singular vectors, singular values, and right
singular vectors, respectively, such that $M=USV^{\top}$. The matrix
square root of $M$, denoted as $\sqrt{M}$, is given by \cite{song2022fast}:
\begin{equation}
	\sqrt{M}=U\sqrt{S}V^{\top}\in\mathbb{R}^{m_{M}\times m_{M}},\label{eq:sqrtm}
\end{equation}
where the square root of a diagonal matrix, such as $S$, is computed
by taking the square root of its diagonal elements. The Unscented
Transform (UT) is an approach used to estimate the probability distribution
of a RV after it undergoes through a nonlinear transformation \cite{julier1997new}.
The problem addressed by the UT involves determining the resultant
distribution of a random variable $D\in\mathbb{R}^{m_{D}}$ after
it is connected through a nonlinear function $\operatorname{f_{nl}}(.)$
to another random variable $C\in\mathbb{R}^{m_{C}}$, where the distribution
of $C$ is known:
\[
D=\operatorname{f_{nl}}(C)
\]
Let $C$ be from a Gaussian distribution with a mean of $\overline{C}$
and a covariance matrix of $P_{C}$. The sigma points representing
$\mathbb{P}(C)$, denoted as the set $\{\mathcal{C}_{j}\}$ are calculated
as:
\begin{equation}
	\left\{ \begin{aligned}\mathcal{C}_{0} & =\overline{C}\in\mathbb{R}^{m_{C}}\\
		\mathcal{C}_{j} & =\overline{C}+\left(\sqrt{(m_{C}+\lambda)P_{C}}\right)_{j}
		\\
		\mathcal{C}_{j+m_{C}} & =\overline{C}-\left(\sqrt{(m_{C}+\lambda)P_{C}}\right)_{j},\hspace{0.3cm}j=\{1,2,\ldots,2m_{C}\}
	\end{aligned}
	\right.\label{eq:default_sigma}
\end{equation}
where $\lambda\in\mathbb{R}$ is a scaling parameter, and $\left(\sqrt{(m_{C}+\lambda)P_{C}}\right)_{j}$
is the $j$the column of matrix square root of $(m_{C}+\lambda)P_{C}$
(as defined in \eqref{eq:sqrtm}) of $(m_{C}+\lambda)P_{C}$. Hence,
from \eqref{eq:default_sigma} each sigma point $\mathcal{C}_{j}$
will be propagated through the nonlinear function and sigma points
representing $\mathbb{P}(D)$, denoted as the set $\{\mathcal{D}_{j}\}$
are calculated as:
\[
\mathcal{D}_{j}=\operatorname{f_{nl}}(\mathcal{C}_{j})\qquad\qquad j=\{0,2,\ldots,2m_{\mathbb{C}}\}
\]
Accordingly, the weighted mean and covariance of the set $\{\mathcal{D}_{j}\}$
provide an accurate representation up to the third degree of the real
mean and covariance matrix of $\mathcal{D}$. The weights can be found
as:
\begin{equation}
	\left\{ \begin{aligned}w_{0}^{m} & =\frac{\lambda}{\lambda+m_{C}}\in\mathbb{R}\\
		w_{0}^{c} & =\frac{\lambda}{\lambda+m_{C}}+1-\alpha^{2}+\beta\in\mathbb{R}\\
		w_{j}^{m} & =w_{j}^{c}=\frac{1}{2(m_{C}+\lambda)}\in\mathbb{R},\hspace{0.3cm}j=\{1,2,\ldots,2m_{C}\}
	\end{aligned}
	\right.\label{eq:default_weights}
\end{equation}
where $\alpha$ and $\beta$ are scaling parameters in $\mathbb{R}$.
Given \eqref{eq:default_weights}, the estimated mean $\hat{\overline{\mathcal{D}}}$
and covariance $\hat{P}_{\mathcal{D}}$ of the distribution are calculated
as follows:
\begin{align}
	\hat{\overline{\mathcal{D}}} & =\sum_{j=0}^{2m_{\mathcal{C}}}w_{j}^{m}\mathcal{D}_{j}\label{eq:meanEstimate}\\
	\hat{P}_{\mathcal{D}} & =\sum_{j=0}^{2m_{\mathcal{C}}}w_{j}^{c}[\mathcal{D}_{j}-\hat{\overline{\mathcal{D}}}][\mathcal{D}_{j}-\hat{\overline{\mathcal{D}}}]^{\top}\in\mathbb{R}^{m_{\mathcal{D}}\times m_{\mathcal{D}}}\label{eq:covEstimate}
\end{align}

\section{Problem Formulation and Measurements\label{sec:Problem-Formulation}}

In this section, we will develop a state transition function that
describes the relationship between the current state, the previous
state, and the input vectors. Additionally, a measurement function
is formulated to describe the relationship between the state vector
and the measurement vector. These functions are crucial for enhancing
filter performance.

\subsection{Model Kinematics}

Consider a rigid-body moving in 3D space, with an angular velocity
$\omega\in\mathbb{R}^{3}$ and an acceleration $a\in\mathbb{R}^{3}$,
measured and expressed w.r.t $\{\mathcal{B}\}$. Let the position
$p\in\mathbb{R}^{3}$ and linear velocity $v\in\mathbb{R}^{3}$ of
the vehicle be expressed in $\{\mathcal{W}\}$, while the vehicle's
orientation in terms of quaternion $q\in\mathbb{S}^{3}$ be expressed
in $\{\mathcal{B}\}$. The system kinematics in continuous space can
then be expressed as \cite{hashim2021geometricNAV,hashim2021gps}:
\begin{equation}
	\left\{ \begin{aligned}\dot{q} & =\frac{1}{2}\Gamma(\omega)q\in\mathbb{S}^{3}\\
		\dot{p} & =v\in\mathbb{R}^{3}\\
		\dot{v} & =g+R_{q}(q)a\in\mathbb{R}^{3}
	\end{aligned}
	\right.\label{c_dyn}
\end{equation}
where
\[
\Gamma(\omega)=\left[\begin{array}{cc}
	0 & -\omega^{\top}\\
	\omega & -[\omega]_{\times}
\end{array}\right]\in\mathbb{R}^{4\times4}
\]
The kinematics expression in \eqref{c_dyn} can be re-arranged in
a geometric form in a similar manner to geometric navigation in \cite{hashim2021geometric,hashim2021gps}
as follows:
\begin{equation}
	\left[\begin{array}{c}
		\dot{q}\\
		\dot{p}\\
		\dot{v}\\
		0
	\end{array}\right]=\underbrace{\left[\begin{array}{cccc}
			\frac{1}{2}\Gamma(\omega)q & 0 & 0 & 0\\
			0 & 0 & I_{3} & 0\\
			0 & 0 & 0 & g+R_{q}(q)a\\
			0 & 0 & 0 & 0
		\end{array}\right]}_{M^{c}(q,\omega,a)}\left[\begin{array}{c}
		q\\
		p\\
		v\\
		1
	\end{array}\right]\label{eq:dyn_cont}
\end{equation}
As the sensors operate in discrete space, the expression in \eqref{eq:dyn_cont}
is discretized for filter derivation and implementation. Let $q_{k}\in\mathbb{S}^{3}$,
$p_{k}\in\mathbb{R}^{3}$, $v_{k}\in\mathbb{R}^{3}$, $\omega_{k}\in\mathbb{R}^{3}$,
and $a_{k}\in\mathbb{R}^{3}$ represent $q\in\mathbb{S}^{3}$, $p\in\mathbb{R}^{3}$,
$v\in\mathbb{R}^{3}$, $\omega\in\mathbb{R}^{3}$, and $a\in\mathbb{R}^{3}$
at the $k$th discrete time step, respectively. The system kinematics
can then be discretized with a sample time of $dT$, in a similar
manner to the approach described in \cite{hashim2021geometric,hashim2021gps}:
\begin{equation}
	\left[\begin{array}{c}
		q_{k}\\
		p_{k}\\
		v_{k}\\
		1
	\end{array}\right]=\exp(M_{k-1}^{c}dT)\left[\begin{array}{c}
		q_{k-1}\\
		p_{k-1}\\
		v_{k-1}\\
		1
	\end{array}\right]\label{eq:dyn_dis}
\end{equation}
where $M_{k-1}^{c}=M^{c}(q_{k-1},\omega_{k-1},a_{k-1})$. At the $k$th
time step, the measured angular velocity $\omega_{m,k}\in\mathbb{R}^{3}$
and linear acceleration $a_{m,k}\in\mathbb{R}^{3}$ are affected by
additive white noise and biases ($b_{\omega,k}\in\mathbb{R}^{3}$
for angular velocity and $b_{a,k}\in\mathbb{R}^{3}$ for linear acceleration),
which are modeled as random walks. The relationship between the true
and measured values can be expressed as: 
\begin{equation}
	\left\{ \begin{aligned}\omega_{m,k} & =\omega_{k}+b_{\omega,k}+n_{\omega,k}\\
		a_{m,k} & =a_{k}+b_{a,k}+n_{a,k}\\
		b_{\omega,k} & =b_{\omega,k-1}+n_{b\omega,k-1}\\
		b_{a,k} & =b_{a,k-1}+n_{ba,k-1}
	\end{aligned}
	\right.\label{a_m}
\end{equation}
where $n_{\omega,k}\in\mathbb{R}^{3}$, $n_{a,k}\in\mathbb{R}^{3}$,
$n_{b\omega,k}\in\mathbb{R}^{3}$, and $n_{ba,k}\in\mathbb{R}^{3}$
are noise vectors from zero mean Gaussian distributions and covariance
matrices of $C_{\omega,k}$, $C_{a,k}$, $C_{b\omega,k}$, and $C_{ba,k}$,
respectively. From \eqref{eq:dyn_dis} and \eqref{a_m}, let us define
the following state vector: 
\begin{equation}
	x_{k}=\begin{bmatrix}q_{k}^{\top} & p_{k}^{\top} & v_{k}^{\top} & b_{\omega,k}^{\top} & b_{a,k}^{\top}\end{bmatrix}^{\top}\in\mathbb{R}^{m_{x}}\label{state}
\end{equation}
with $m_{x}=16$ representing the state dimension. Let the augmented
and additive noise vectors with dimensions of $m_{n_{x}}=6$ and $m_{n_{w}}=16$
be defined as: 
\begin{equation}
	\left\{ \begin{aligned}n_{x,k} & =\begin{bmatrix}n_{\omega,k}^{\top} & n_{a,k}^{\top}\end{bmatrix}^{\top}\in\mathbb{R}^{m_{n_{x}}}\\
		n_{w,k} & =\begin{bmatrix}0_{10\times1}^{\top} & n_{b\omega,k}^{\top} & n_{ba,k}^{\top}\end{bmatrix}^{\top}\in\mathbb{R}^{m_{n_{w}}}
	\end{aligned}
	\right.\label{eq:noise vector}
\end{equation}
Let the input vector $u_{k}$ at time step $k$ be defined as: 
\begin{equation}
	u_{k}=\begin{bmatrix}\omega_{m,k}^{\top} & a_{m,k}^{\top}\end{bmatrix}^{\top}\in\mathbb{R}^{m_{u}}\label{eq:input vector}
\end{equation}
where $m_{u}=6$. From \eqref{state} and \eqref{eq:noise vector},
let us define the augmented state vector as:
\begin{equation}
	x_{k}^{a}=\begin{bmatrix}x_{k}^{\top} & n_{x,k}^{\top}\end{bmatrix}^{\top}\in\mathbb{R}^{m_{a}}\label{eq:augmented state vector}
\end{equation}
with $m_{a}=m_{x}+m_{n_{x}}$. Combining \eqref{eq:dyn_dis}, \eqref{a_m},
\eqref{eq:noise vector}, \eqref{eq:input vector}, and \eqref{eq:augmented state vector},
one can re-express the system kinematics in discrete space as follows:
\begin{equation}
	x_{k}=\operatorname{f}(x_{k-1}^{a},u_{k-1})+n_{w,k-1}\label{state transition}
\end{equation}
where $\operatorname{f}(\cdot):\mathbb{R}^{m_{a}}\times\mathbb{R}^{m_{u}}\rightarrow\mathbb{R}^{m_{x}}$
denotes the state transition matrix.

\subsection{VIN Measurement Model}

\begin{figure}[t]
	\centering \includegraphics[scale=0.23]{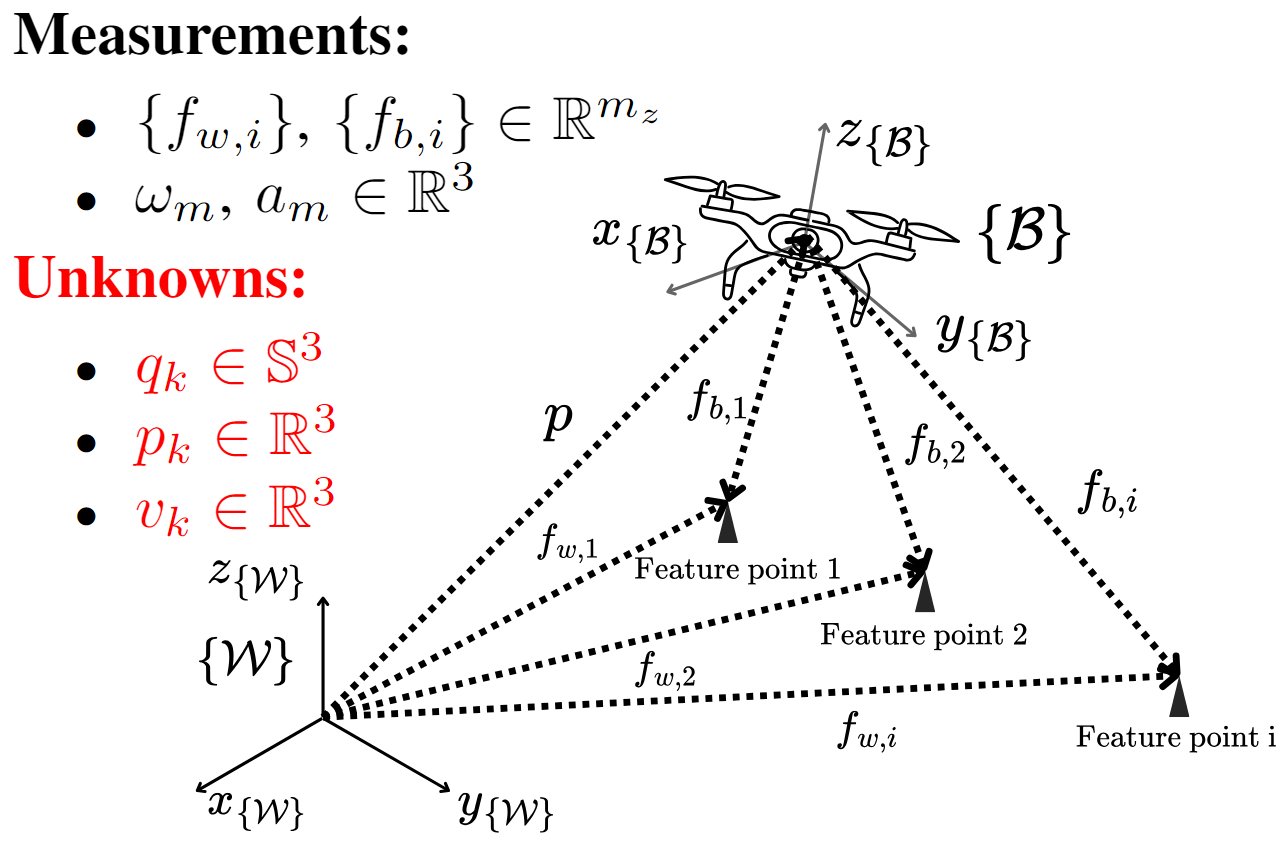} \caption{Navigation problem visualization in the 6 DoF.}
	\label{fig:Schematics}
\end{figure}

Consider $f_{w,i}\in\mathbb{R}^{3}$ to be the coordinates of the
$i$th feature point in $\{\mathcal{W}\}$ obtained by using a series
of stereo camera observations and $f_{b,i}\in\mathbb{R}^{3}$ be the
coordinates of the $i$th feature point in $\{\mathcal{B}\}$ reconstructed
using the stereo camera measurement at the $k$th time step. The model
describing the relationship between each $f_{w,i}$ and $f_{b,i}$
can be expressed as \cite{hashim2021geometricNAV,wei2022optimization}:
\begin{align}
	f_{b,i}=R_{q}(q_{k})^{\top}(f_{w,i}-p_{k})+n_{f,i} & \in\mathbb{R}^{3}\label{W2B}
\end{align}
where $n_{f,i}$ refers to white Gaussian noise associated with each
measurement for all $i\in\{1,2,\ldots,m_{f}\}$ and $m_{f}\in\mathbb{R}$
denotes the number of feature points detected in each measurement
step. Note that the quantity of feature points $m_{f}$ may vary across
images obtained at different measurement steps. Let us define the
set of feature points in $\{\mathcal{W}\}$ as $f_{w}$, the set of
feature points in $\{\mathcal{B}\}$ as $f_{b}$, and the measurement
noise vector as $n_{f}$ such that
\begin{equation}
	\left\{ \begin{aligned}f_{b} & =\begin{bmatrix}f_{b,1}^{\top} & f_{b,2}^{\top} & \cdots & f_{b,m_{f}}^{\top}\end{bmatrix}^{\top}\in\mathbb{R}^{3m_{f}}\\
		f_{w} & =\begin{bmatrix}f_{w,1}^{\top} & f_{w,2}^{\top} & \cdots & f_{w,m_{f}}^{\top}\end{bmatrix}^{\top}\in\mathbb{R}^{3m_{f}}\\
		n_{f} & =\begin{bmatrix}n_{f,1}^{\top} & n_{f,2}^{\top} & \cdots & n_{f,m_{f}}^{\top}\end{bmatrix}^{\top}\in\mathbb{R}^{3m_{f}}
	\end{aligned}
	\right.\label{eq:feature points}
\end{equation}
Using \eqref{eq:feature points}, the term in \eqref{W2B} can be
rewritten in form of a measurement function as follows:
\begin{equation}
	f_{b,k}=z_{k}=\operatorname{h}(x_{k},f_{w})+n_{f,k}\in\mathbb{R}^{m_{z}}\label{eq:measurement}
\end{equation}
with
\[
n_{f}\sim N(0,C_{f}\in\mathbb{R}^{m_{z}\times m_{z}})
\]
Note that $z_{k}$ is the measurement vector at the $k$th time step
and $m_{z}=3m_{f}\in\mathbb{Z}^{+}$ is the dimension of the measurement
vector. Typically, the camera sensor operates at a lower sampling
frequency when compared to the IMU sensor. As a result, there may
be a timing discrepancy between the availability of images and the
corresponding IMU data. The conceptual illustration of the VIN problem
is presented in Fig. \ref{fig:Schematics}.

\section{QNUKF-based VIN Design \label{sec:Filter}}

To develop the QNUKF, we will explore how the Conventional UKF (CUKF)
\cite{ref:ukf} functions within the system. Next, we will modify
each step to address the CUKF limitations. Stochastic filters, such
as UKF models system parameters as RVs and aim to estimate the expected
values of their distributions, which are then used as the estimates
of the parameters. Let $X_{k}$ and $Z_{k}$ represent the RVs associated
with the real values of $x_{k}$ and $z_{k}$, respectively. In this
model, the expected values of the RVs are accounted as the real values.
However, it should be noted that directly calculating these expected
values is not feasible; instead, they are estimated. Let the estimated
expected values of the probabilities $\mathbb{P}(X_{k-1}|Z_{k-1})$
and $\mathbb{P}(X_{k}|Z_{k-1})$ be $\hat{x}_{k-1|k-1}$ and $\hat{x}_{k|k-1}$,
respectively, and the expected value of $\mathbb{P}(Z_{k})$ be $\hat{z}_{k|k-1}$.

\subsection{Filter Initialization}

\subsubsection*{Step 1. State and Covariance Initialization}

At the QNUKF initialization, we establish the initial state estimate
$\hat{x}_{0|0}$ and the covariance matrix $P_{0|0}$. While this
step is standard in CUKF, managing quaternion-based navigation introduces
unique challenges related to quaternion representation and covariance.
A critical challenge arises from the fact that the covariance is computed
based on differences from the mean. Thereby, straightforward quaternion
subtraction is not feasible. To address this limitation, we employ
the custom quaternion subtraction introduced in \eqref{eq:q-q} and
specifically designed for quaternion operations. Moreover, quaternions
inherently possess three degrees of freedom while having four components,
requiring adjustments to the covariance matrix to reflect the reduced
dimensionality. Let the initial estimated state vector be defined
as $\hat{x}_{0|0}=\left[\hat{x}_{0|0,q}^{\top},\hat{x}_{0|0,-}^{\top}\right]^{\top}\in\mathbb{R}^{m_{x}}$
(see \eqref{state}), where $\hat{x}_{0|0,q}\in\mathbb{S}^{3}$ and
$\hat{x}_{0|0,-}\in\mathbb{R}^{m_{x}-4}$ represent the quaternion
and non-quaternion components of $\hat{x}_{0|0}$, respectively, with
$m_{x}$ being the number of rows in the state vector. The filter
is initialized as:
\begin{align}
	\hat{x}_{0|0} & =\left[\hat{x}_{0|0,q}^{\top},\hat{x}_{0|0,-}^{\top}\right]^{\top}\in\mathbb{R}^{m_{x}}\label{eq:QNUKF_init1}\\
	P_{0|0} & =\text{diag}(P_{\hat{x}_{0|0,q}},P_{0|0,-})\in\mathbb{R}^{(m_{x}-1)\times(m_{x}-1)}\label{eq:QNUKF_init2}
\end{align}
where $P_{0|0,q}\in\mathbb{R}^{3\times3}$, $P_{0|0,-}\in\mathbb{R}^{(m_{x}-4)\times(m_{x}-4)}$,
and $P_{0|0}$ represent the covariance matrices corresponding with
the uncertainties of $\hat{x}_{0|0,q}$, $\hat{x}_{0|0,-}$, and $\hat{x}_{0|0}$,
respectively.

\subsection{Prediction}

The prediction step involves estimating the state of the system at
the next time step leveraging the system's kinematics given the initial
state \eqref{eq:QNUKF_init1} and covariance \eqref{eq:QNUKF_init2}.
At each $k$ time step, the process begins with augmenting the state
vector with non-additive noise terms, a critical step that ensures
the filter accounts for the uncertainty which these noise terms introduce.
Subsequently, sigma points are generated from the augmented state
vector and covariance matrix. These sigma points effectively represent
the distribution $\mathbb{P}(X_{k-1}|Z_{k-1})$, capturing the state's
uncertainty at the previous time step given the previous observation,
see \eqref{eq:default_sigma}. \eqref{eq:AB}. The next phase involves
computing the posterior sigma points by applying the state transition
function to each sigma point, thereby representing the distribution
$\mathbb{P}(X_{k}|X_{k-1})$ which is equivalent to $\mathbb{P}(X_{k}|Z_{k-1})$.
This distribution reflects the predicted state's uncertainty before
the current observation is considered. The predicted state vector
is then obtained as the weighted average of these posterior sigma
points. This process is described in detail in this subsection. 

\subsubsection*{Step 2. Augmentation}

Prior to calculating the sigma points, the state vector and covariance
matrix are augmented to incorporate non-additive process noise. This
involves adding rows and columns to the state vector and covariance
matrix to represent the process noise. The expected value of the noise
terms, zero in this case, are augmented to the previous state estimate
$\hat{x}_{k-1|k-1}$ to form the augmented state vector $\hat{x}_{k-1|k-1}^{a}$.
Similarly, the last estimated state covariance matrix $P_{k-1|k-1}$
and the noise covariance matrix $C_{x,k}$ are augmented together
to form the augmented state covariance matrix $P_{k-1|k-1}^{a}$.
This is formulated as follows:
\begin{align}
	\hat{x}_{k-1|k-1}^{a} & =\left[\hat{x}_{k-1|k-1}^{\top},0_{m_{n_{x}}\times1}^{\top}\right]^{\top}\in\mathbb{R}^{m_{a}}\label{eq:QNUKF_augment1}\\
	P_{k-1|k-1}^{a} & =\text{diag}(P_{k-1|k-1},C_{x,k})\in\mathbb{R}^{(m_{a}-1)\times(m_{a}-1)}\label{eq:QNUKF_augment2}
\end{align}
The covariance matrix $C_{x,k}$ in \eqref{eq:QNUKF_augment2} is
defined by:
\begin{equation}
	C_{x,k}=\text{diag}(C_{\omega,k},C_{a,k})\in\mathbb{R}^{m_{n_{x}}\times m_{n_{x}}}\label{eq:C_nx}
\end{equation}

\subsubsection*{Step 3. Sigma Points Calculation}

Conventionally, sigma points are computed based on the augmented state
estimate and covariance matrix. They serve as a representation for
the distribution and uncertainty of $\mathbb{P}(x_{k-1|k-1}|z_{k-1})$.
In other words, the greater the uncertainty of the last estimate reflected
by a larger last estimated covariance, the more widespread the sigma
points will be around the last estimated state. Reformulating \eqref{eq:default_sigma}
according to the CUKF algorithm, the sigma points, crucial for propagating
the state through the system dynamics, are computed using the following
equation:
\begin{equation}
	\left\{ \begin{aligned}\mathcal{X}_{k-1|k-1,0}^{a} & =\hat{x}_{k-1|k-1}^{a}\in\mathbb{R}^{m_{a}}\\
		\mathcal{X}_{k-1|k-1,j}^{a} & =\hat{x}_{k-1|k-1}^{a}+\left(\sqrt{(m_{a}+\lambda)P_{k-1|k-1}^{a}}\right)_{j}
		\\
		\mathcal{X}_{k-1|k-1,j+m_{a}}^{a} & =\hat{x}_{k-1|k-1}^{a}-\left(\sqrt{(m_{a}+\lambda)P_{k-1|k-1}^{a}}\right)_{j},\\
		& \hspace{2cm}j=\{1,2,\ldots,2m_{a}\}
	\end{aligned}
	\right.\label{eq:Sigma_CUKF}
\end{equation}
Challenges arise when using the sigma points calculation as described
in \eqref{eq:Sigma_CUKF} for quaternion-based navigation problems
due to the differing mathematical spaces involved. For the sake of
brevity, let us define $\delta\hat{x}_{k-1,j}^{a}:=\left(\sqrt{(m_{a}+\lambda)P_{k-1|k-1}^{a}}\right)_{j}\in\mathbb{R}^{m_{a}-1}$.
It is possible to divide $\hat{x}_{k-1|k-1}^{a}$ and $\delta\hat{x}_{k-1,j}^{a}$
into their attitude and non-attitude parts: $\hat{x}_{k-1|k-1,q}\in\mathbb{S}^{3}$,
$\delta\hat{x}_{k-1,j,r}^{a}\in\mathbb{R}^{3}$, and $\hat{x}_{k-1|k-1,-}^{a}\in\mathbb{R}^{m_{a}-4}$,
$\delta\hat{x}_{k-1,j,-}^{a}\in\mathbb{R}^{m_{a}-4}$, as outlined
below:
\begin{equation}
	\left\{ \begin{aligned}\hat{x}_{k-1|k-1}^{a} & =\left[(\hat{x}_{k-1|k-1,q}^{a})^{\top},(\hat{x}_{k-1|k-1,-}^{a})^{\top}\right]^{\top}\\
		\delta\hat{x}_{k-1,j}^{a} & =\left[(\delta\hat{x}_{k-1,j,r}^{a})^{\top},(\delta\hat{x}_{k-1,j,-}^{a})^{\top}\right]^{\top}
	\end{aligned}
	\right.\label{eq:693}
\end{equation}
From \eqref{eq:693}, given that $\delta\hat{x}_{k-1,j,r}^{a}$ is
a rotation vector in $\mathbb{R}^{3}$ and $\hat{x}_{k-1|k-1,q}^{a}$
is a quaternion expression in $\mathbb{S}^{3}$, direct addition or
subtraction is not feasible. To resolve this limitation, we employ
the custom subtraction \eqref{eq:q-r} and addition \eqref{eq:q+r}
operators, as previously defined. The operation map $\mathbb{S}^{3}\times\mathbb{R}^{3}\rightarrow\mathbb{S}^{3}$
effectively manages the interaction between quaternions and rotation
vectors. Additionally, the quaternion-based navigation problem inherently
possesses $m_{a}-1$ degrees of freedom due to the reduced dimensionality
of the quaternion. To resolve this limitation, the propose QNUKF modifies
$\delta\hat{x}_{k-1,j}^{a}$ as follows:
\begin{equation}
	\delta\hat{x}_{k-1,j}^{a}:=\left(\sqrt{(m_{a}-1+\lambda)P_{k-1|k-1}^{a}}\right)_{j}\in\mathbb{R}^{m_{a}-1}\label{eq:del_k-1_QNUKF}
\end{equation}
Overall, considering \eqref{eq:693} and \eqref{eq:del_k-1_QNUKF},
the QNUKF utilizes the modified version of \eqref{eq:Sigma_CUKF}
as follows:
\begin{equation}
	\left\{ \begin{aligned}\mathcal{X}_{k-1|k-1,0}^{a} & =\hat{x}_{k-1|k-1}^{a}\in\mathbb{R}^{m_{a}}\\
		\mathcal{X}_{k-1|k-1,j}^{a} & =\hat{x}_{k-1|k-1}^{a}\oplus\delta\hat{x}_{j}^{a}\\
		& :=\begin{bmatrix}\hat{x}_{k-1|k-1,q}^{a}\oplus\delta\hat{x}_{k-1,j,r}^{a}\\
			\hat{x}_{k-1|k-1,-}^{a}+\delta\hat{x}_{k-1,j,-}^{a}
		\end{bmatrix}\in\mathbb{R}^{m_{a}}\\
		\mathcal{X}_{k-1|k-1,j+m_{a}}^{a} & =\hat{x}_{k-1|k-1}^{a}\ominus\delta\hat{x}_{j}^{a}\\
		& :=\begin{bmatrix}\hat{x}_{k-1|k-1,q}^{a}\ominus\delta\hat{x}_{k-1,j,r}^{a}\\
			\hat{x}_{k-1|k-1,-}^{a}-\delta\hat{x}_{k-1,j,-}^{a}
		\end{bmatrix}\in\mathbb{R}^{m_{a}},\\
		& \hspace{1.5cm}j=\{1,2,\ldots,2(m_{a}-1)\}
	\end{aligned}
	\right.\label{eq:Sigma_QNUKF}
\end{equation}

\begin{figure*}[!t]
	\centering \includegraphics[width=1.8\columnwidth]{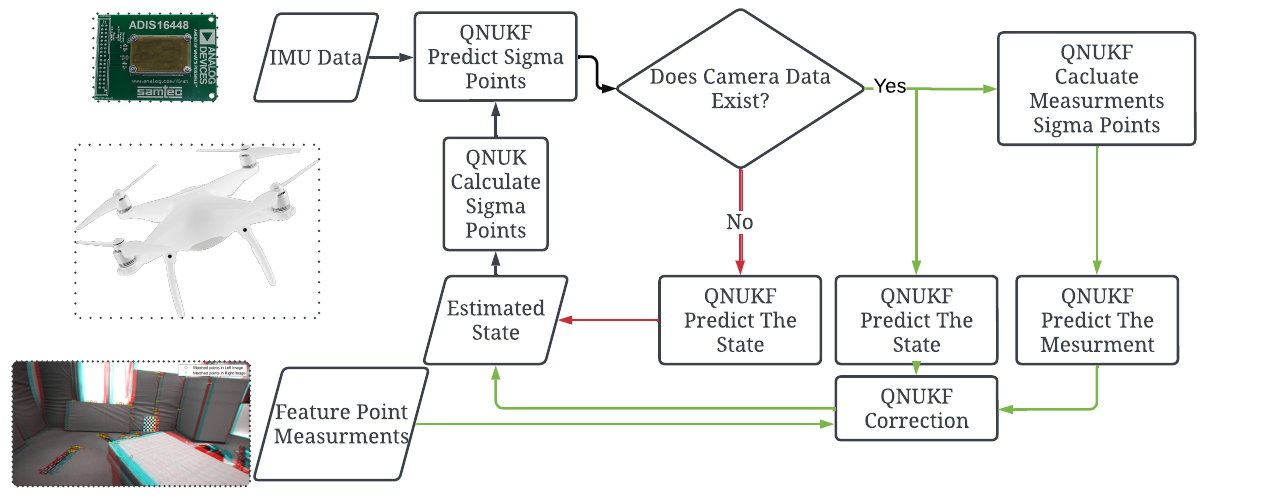}
	\caption{Schematic diagram of the proposed QNUKF. First, the sigma points based
		on the last estimated state vector and its covariance matrix are calculated.
		Using these sigma points and IMU measurements, the prediction step
		is performed. If camera data exists at that sample point, the correction
		(also known as update) step is performed. The state and covariance
		estimation at that time step is the result of the correction step
		if camera data exists, and the prediction step if it does not.}
	\label{fig:flowchart}
\end{figure*}

\subsubsection*{Step 4. Propagate Sigma Points}

The sigma points $\{\mathcal{X}_{k-1|k-1,j}^{a}\}$ are propagated
through the state transition function, as defined in \eqref{state transition},
to calculate the propagated sigma points $\{\mathcal{X}_{k|k-1,j}\}$.
These points represent the probability distribution $\mathbb{P}(X_{k}|X_{k-1})$,
and are computed as follows:
\begin{align}
	\mathcal{X}_{k|k-1,j}= & \operatorname{f}(\mathcal{X}_{k-1|k-1,j}^{a},u_{k-1})\in\mathbb{R}^{m_{x}}\label{eq:propagate_sigma_points_augmented}
\end{align}
for all $j=\{0,1,2,\ldots,2(m_{a}-1)\}$.

\subsubsection*{Step 5. Compute Predicted Mean and Covariance}

The predicted mean $\hat{x}_{k|k-1}$ and covariance $P_{k|k-1}$
are conventionally computed using the predicted sigma points by reformulating
\eqref{eq:meanEstimate} and \eqref{eq:covEstimate} as:
\begin{align}
	\hat{x}_{k|k-1} & =\sum_{j=0}^{2m_{a}}w_{j}^{m}\mathcal{X}_{k|k-1,j}\in\mathbb{R}^{m_{x}}\label{eq:state mean - CUKF}\\
	P_{k|k-1} & =\sum_{j=0}^{2m_{a}}\left[w_{j}^{c}(\mathcal{X}_{k|k-1,j}-\hat{x}_{k|k-1})(\mathcal{X}_{k|k-1,j}-\hat{x}_{k|k-1})^{\top}\right]\nonumber \\
	& \qquad+C_{w,k}\in\mathbb{R}^{(m_{x}-1)\times(m_{x}-1)}\label{eq:state cov - CUKF}
\end{align}
where $w_{j}^{m}$ and $w_{j}^{c}\in\mathbb{R}$ represent the weights
associated with the sigma points and are found by \eqref{eq:default_weights}.
Additionally, $C_{w,k}\in\mathbb{R}^{m_{x}-1}$ denotes the process
noise covariance matrix, which is the covariance matrix of the noise
vector $n_{w,k}$ as defined in \eqref{eq:noise vector}. Note that
the reduced dimensionality of $C_{w,k}$, similar to $P_{k|k-1}$
and $P_{0|0}$, is due to the covariance matrices of quaternion random
variables being described by $3\times3$ matrices, reflecting the
three degrees of freedom inherent to quaternions. $C_{w,k}$ is defined
as
\begin{equation}
	C_{w,k}=\begin{bmatrix}0_{9\times9} & 0_{3\times3} & 0_{3\times3}\\
		0_{9\times9} & C_{b\omega,k} & 0_{3\times3}\\
		0_{9\times9} & 0_{3\times3} & C_{ba,k}
	\end{bmatrix}\in\mathbb{R}^{(m_{x}-1)\times(m_{x}-1)}\label{eq:Q}
\end{equation}
To tailor \eqref{eq:state mean - CUKF}, \eqref{eq:state cov - CUKF},
and \eqref{eq:default_weights} for the quaternion-based navigation
problem, the following modifications are necessary: (1) The QNUKF
utilizes $m_{a}-1$ sigma points, therefore, all occurrences of $m_{a}$
in \eqref{eq:state mean - CUKF}, \eqref{eq:state cov - CUKF}, and
\eqref{eq:default_weights} should be replaced with $m_{a}-1$; (2)
The average of quaternion components of the sigma points in \eqref{eq:state mean - CUKF}
should be computed using the quaternion weighted average method described
in \eqref{eq:weighted average}. This is necessary since the typical
vector averaging method in \eqref{eq:state mean - CUKF} cannot be
applied directly to quaternions; (3) The subtraction of quaternion
components of the sigma point vector by the quaternion components
of the estimated mean in \eqref{eq:state cov - CUKF} should utilize
the custom subtraction method defined in \eqref{eq:q-q}. This operation
maps $\mathbb{S}^{3}\times\mathbb{S}^{3}$ to $\mathbb{R}^{3}$, reflecting
the fact that quaternions, which exist in $\mathbb{S}^{3}$ space,
cannot be subtracted using conventional vector subtraction techniques.

To facilitate these operations, let us divide $\hat{x}_{k|k-1}$ and
$\mathcal{X}_{k|k-1,j}$ to their quaternion ($\hat{x}_{k|k-1,q}\in\mathbb{S}^{3}$,
$\mathcal{X}_{k|k-1,j,q}\in\mathbb{S}^{3}$) and non-quaternion ($\hat{x}_{k|k-1,-}\in\mathbb{R}^{n_{x}-4}$,
$\mathcal{X}_{k|k-1,j,-}\in\mathbb{R}^{m_{x}-4}$) components. These
divisions can be represented as follows:
\begin{align}
	\hat{x}_{k|k-1} & =\begin{bmatrix}\hat{x}_{k|k-1,q}^{\top} & \hat{x}_{k|k-1,-}^{\top}\end{bmatrix}^{\top}\mathbb{R}^{m_{x}}\label{eq:x_k_k_1}\\
	\mathcal{X}_{k|k-1,j} & =\begin{bmatrix}\mathcal{X}_{k|k-1,q}^{\top} & \mathcal{X}_{k|k-1,-}^{\top}\end{bmatrix}^{\top}\in\mathbb{R}^{m_{x}}\label{eq:QNUKF_Xkk-1}
\end{align}
Consequently, equations \eqref{eq:state mean - CUKF}, \eqref{eq:state cov - CUKF},
and \eqref{eq:default_weights} can be reformulated to accommodate
these quaternion-specific processes, utilizing the divisions formulated
in \eqref{eq:x_k_k_1} and \eqref{eq:QNUKF_Xkk-1} as follows:
\begin{align}
	\hat{x}_{k|k-1} & =\begin{bmatrix}\operatorname{QWA}(\{\mathcal{X}_{k|k-1,j,q}\},\{w_{j}^{m}\})\\
		{\displaystyle \sum_{j=0}^{2(m_{a}-1)}w_{j}^{m}\mathcal{X}_{k|k-1,j,-}}
	\end{bmatrix}\in\mathbb{R}^{m_{x}}\label{eq:state mean - QNUKF}
\end{align}
{\small 
	\begin{align}
		P_{k|k-1} & =\sum_{j=0}^{2(m_{a}-1)}\left[w_{j}^{c}(\mathcal{X}_{k|k-1,j}\ominus\hat{x}_{k|k-1})(\mathcal{X}_{k|k-1,j}\ominus\hat{x}_{k|k-1})^{\top}\right]\nonumber \\
		& \qquad+C_{w,k}\in\mathbb{R}^{(m_{x}-1)\times(m_{x}-1)}\label{eq:state cov - QNUKF}
	\end{align}
}and
\begin{equation}
	\left\{ \begin{aligned}w_{0}^{m} & =\frac{\lambda}{\lambda+(m_{a}-1)}\in\mathbb{R}\\
		w_{0}^{c} & =\frac{\lambda}{\lambda+(m_{a}-1)}+1-\alpha^{2}+\beta\in\mathbb{R}\\
		w_{j}^{m} & =w_{j}^{c}=\frac{1}{2((m_{a}-1)+\lambda)}\in\mathbb{R}\\
		& \hspace{2cm}j=\{1,\ldots,2(m_{a}-1)\}
	\end{aligned}
	\right.\label{eq:weights-QNUKF}
\end{equation}
where
\begin{align}
	\mathcal{X}_{k|k-1,j}\ominus\hat{x}_{k|k-1} & =\begin{bmatrix}\mathcal{X}_{k|k-1,j,q}\ominus\hat{x}_{k|k-1,q}\\
		\mathcal{X}_{k|k-1,j,-}-\hat{x}_{k|k-1,-}
	\end{bmatrix}\in\mathbb{R}^{m_{x}-1}\label{eq:x-x}
\end{align}

\subsection{Update}

In the update step, the propagated sigma points are first processed
through the measurement function to generate the measurement sigma
points, which represent $\mathbb{P}(Z_{k}|X_{k})$. Subsequently,
utilizing $\mathbb{P}(X_{k-1}|Z_{k-1})$, $\mathbb{P}(X_{k}|X_{k-1})$,
and $\mathbb{P}(Z_{k}|X_{k})$, the conditional probability $\mathbb{P}(X_{k}|Z_{k})$
is estimated. The expected value derived from $\mathbb{P}(X_{k}|Z_{k})$
will serve as the state estimate for the $k$th step.

\subsubsection*{Step 6. Predict Measurement and Calculate Covariance}

In this step, each propagated sigma point $\mathcal{X}_{k|k-1,j}$,
as found in \eqref{eq:propagate_sigma_points_augmented}, is processed
through the measurement function defined in \eqref{eq:measurement}
to compute the $j$th predicted sigma point $\mathcal{Z}_{k|k-1,j}$.
The set $\{\mathcal{Z}_{k|k-1,j}\}$ represent the probability distribution
$\mathbb{P}(Z_{k}|X_{k})$. For all $j=\{0,1,\ldots,2(m_{a}-1)\}$,
the computation for each sigma point is given by:
\begin{align}
	\mathcal{Z}_{k|k-1,j} & =h(\mathcal{X}_{k|k-1,j},f_{w})\in\mathbb{R}^{m_{z}}\label{eq:propagate measurment}
\end{align}
The expected value of $\mathbb{P}(Z_{k}|X_{k})$ is the maximum likelihood
estimate of the measurement vector. This estimate $\hat{z}_{k|k-1}$
is calculated according to \eqref{eq:zhat - QNUKF}. Additionally,
the estimated measurement covariance matrix $P_{z_{k},z_{k}}\in\mathbb{R}^{m_{z}\times m_{z}}$
and the state-measurement covariance matrix $P_{x_{k},z_{k}}\in\mathbb{R}^{(m_{x}-1)\times m_{z}}$
are computed using the following equations:
\begin{align}
	\hat{z}_{k|k-1} & =\sum_{j=0}^{2(m_{a}-1)}w_{j}^{m}\mathcal{Z}_{k|k-1,j}\label{eq:zhat - QNUKF}\\
	P_{z_{k},z_{k}} & =\sum_{j=0}^{2(m_{a}-1)}w_{j}^{c}[\mathcal{Z}_{k|k-1,j}-\hat{z}_{k|k-1}][\mathcal{Z}_{k|k-1,j}-\hat{z}_{k|k-1}]^{\top}\nonumber \\
	& \hspace{1.5cm}+C_{f}\label{eq:P_zz - QUKF}\\
	P_{x_{k},z_{k}} & =\sum_{j=0}^{2(m_{a}-1)}w_{j}^{c}[\mathcal{X}_{k|k-1,j}\ominus\hat{x}_{k|k-1}][\mathcal{Z}_{k|k-1,j}-\hat{z}_{k|k-1}]^{\top}\label{eq:P_xz - QUKF}
\end{align}
Note that the subtraction operator $\ominus$ in \eqref{eq:P_xz - QUKF}
denotes state difference operator and follows \eqref{eq:x-x}. Now,
the joint distribution of $(X_{k},Z_{k})$ is estimated as follows:
\begin{equation}
	\begin{pmatrix}X_{k}\\[0.5em]
		Z_{k}
	\end{pmatrix}\sim\mathcal{N}\left(\begin{pmatrix}\hat{x}_{k|k-1}\\[0.5em]
		\hat{z}_{k|k-1}
	\end{pmatrix},\begin{pmatrix}P_{k|k-1} & P_{x_{k},z_{k}}\\[0.5em]
		P_{x_{k},z_{k}}^{\top} & P_{z_{k},z_{k}}
	\end{pmatrix}\right)\label{eq:Xk_Zk}
\end{equation}
These covariance calculations are necessary to estimate the probability
distribution $\mathbb{P}(X_{k}|Z_{k})$ in the next step.

\subsubsection*{Step 7. Update State Estimate}

In view of \eqref{eq:Xk_Zk} to \eqref{eq:AB-normal}, let us reformulate
\eqref{eq:AB} to find the mean and covariance estimate of the $\mathbb{P}(X_{k}|Z_{k})$
distribution. It is worth noting that the estimated mean is the estimated
state vector at this time step. This is achieved by computing the
Kalman gain $K_{k}\in\mathbb{R}^{(m_{x}-1)\times m_{z}}$ given the
matrices found in \eqref{eq:P_zz - QUKF} and \eqref{eq:P_xz - QUKF}.
\begin{equation}
	K_{k}=P_{x_{k},z_{k}}P_{z_{k},z_{k}}^{\top}\in\mathbb{R}^{(m_{x}-1)\times m_{z}}\label{eq:K}
\end{equation}
To find the updated covariance estimate of the state distribution
$P_{k|k}$ in the QNUKF, the conventional formula provided in \eqref{eq:update:P}
is utilized:
\begin{equation}
	P_{k|k}=P_{k|k-1}-K_{k}P_{z_{k},z_{k}}K_{k}^{\top}\label{eq:update:P}
\end{equation}
However, computing the updated estimated state vector $\hat{x}_{k|k}$
requires careful consideration. Define the correction vector $\delta\hat{x}_{k|k-1}$
as:
\begin{equation}
	\delta\hat{x}_{k|k-1}:=K_{k}(z_{k}-\hat{z}_{k|k-1})\in\mathbb{R}^{m_{x}-1}\label{eq:correction vector}
\end{equation}
Recall that $z_{k}$ refers to camera feature measurements in $\{\mathcal{B}\}$.
Conventionally, the correction vector calculated in \eqref{eq:correction vector},
is added to the predicted state vector from \eqref{eq:state mean - QNUKF}
to update the state. However, for a quaternion-based navigation problem,
the dimensions of $\delta\hat{x}_{k|k-1}$ and $\hat{x}_{k|k-1}$
do not match, with the former being $m_{x}-1$ and the latter $m_{x}$.
To address this, let us divide $\delta\hat{x}_{k|k-1}$ into its attitude
($\delta\hat{x}_{k|k-1,r}\in\mathbb{R}^{3}$) and non-attitude ($\delta\hat{x}_{k|k-1,-}\in\mathbb{R}^{m_{\star}-4}$)
components:
\begin{equation}
	\delta\hat{x}_{k|k-1}=\begin{bmatrix}\delta\hat{x}_{k|k-1,r}^{\top} & \delta\hat{x}_{k|k-1,-}^{\top}\end{bmatrix}^{\top}\label{eq:delx_k_k_1}
\end{equation}
From \eqref{eq:delx_k_k_1} with \eqref{eq:x_k_k_1}, we observe that
the non-quaternion components of $\hat{x}_{k|k-1}$ and $\delta\hat{x}_{k|k-1}$
are dimensionally consistent and can be combined using the conventional
addition operator. However, the orientation components of $\hat{x}_{k|k-1}$
and $\delta\hat{x}_{k|k-1}$ do not match dimensionally. The former
is represented by a quaternion within the $\mathbb{S}^{3}$ space,
while the latter is a rotation vector in the three-dimensional Euclidean
space $\mathbb{R}^{3}$. To add the orientation parts of the predicted
state and correction vectors, the custom summation operator $\mathbb{S}^{3}\times\mathbb{R}^{3}\rightarrow\mathbb{S}^{3}$
described in \eqref{eq:q+r} should be used as follows
\begin{equation}
	\hat{x}_{k|k}=\hat{x}_{k|k-1}\oplus\delta\hat{x}_{k|k-1}\label{eq:final update - QUNKF}
\end{equation}
where
\begin{equation}
	\hat{x}_{k|k-1}\oplus\delta\hat{x}_{k|k-1}=\begin{bmatrix}\hat{x}_{k|k-1,q}\oplus\delta\hat{x}_{k|k-1,r}\\
		\hat{x}_{k|k-1,-}+\delta\hat{x}_{k|k-1,-}
	\end{bmatrix}\label{eq:QNUKF_correction}
\end{equation}

\subsubsection*{Step 8. Iterate}

Go back to Step 2. and iterate from $k\rightarrow k+1$.

The proposed QNUKF algorithm is visually depicted in Fig. \ref{fig:flowchart}.
The implementation steps of QNUKF algorithm is summarized and outlined
in Algorithm \ref{alg:Alg_QNUKF}.

\begin{rem}\label{rem:The-time-complexities}The time complexities
	of the vanilla EKF and UKF algorithms are $O(m_{x}^{3})$ and $O(m_{a}^{3})$,
	respectively \cite{5152793}. For navigation tasks, these complexities
	reduce to $O((m_{x}-1)^{3})$ and $O((m_{a}-1)^{3})$, respectively,
	due to the reduced dimensionality of the quaternion in the state vector.
	The custom operations proposed in QNUKF, such as the one used in \eqref{eq:QNUKF_correction},
	all have complexities less than $O((m_{a}-1))$ and occur only once
	for each sigma point. Therefore, QNUKF retains the same overall complexity
	as any UKF, which is $O((m_{a}-1)^{3})$. The difference between $m_{a}$
	and $m_{x}$ arises because UKF can capture non-additive noise terms
	by augmenting them into the state vector, whereas EKF linearizes them
	into an additive form.\end{rem}

\begin{algorithm}[!h]
	\caption{\label{alg:Alg_QNUKF}Quaternion Navigation Unscented Kalman Filter}
	
	\textbf{Initialization}:
	\begin{enumerate}
		\item[{\footnotesize{}1:}] Set $\hat{x}_{0|0}\in\mathbb{R}^{m_{x}-4}$, see \eqref{state},
		and $P_{0|0}\in\mathbb{R}^{(m_{x}-1)\times(m_{x}-1)}$. 
		\item[{\footnotesize{}2:}] Set $k=0$ and select the filter design parameters $\lambda$, $\alpha$,
		$\beta\in\mathbb{R}$, along with the noise covariance matrices $C_{a,k}$,
		$C_{\omega,k}$, $C_{ba,k}$, $C_{b\omega,k}\in\mathbb{R}^{3\times3}$,
		and the noise variance $c_{f}\in\mathbb{R}$.
		\item[{\footnotesize{}3:}] Calculate $C_{x,k}$, $C_{w,k}$, and $C_{f}$ as expressed in \eqref{eq:C_nx},
		\eqref{eq:Q}, and \eqref{eq:C_f}, respectively.
	\end{enumerate}
	\textbf{while IMU data exists}
	\begin{enumerate}
		\item[] /{*} Prediction {*}/
		\item[{\footnotesize{}4:}] $\begin{cases}
			\hat{x}_{k-1|k-1}^{a} & =\left[\hat{x}_{k-1|k-1}^{\top},0_{6\times1}^{\top}\right]^{\top}\in\mathbb{R}^{(m_{a}-1)\times1}\\
			P_{k-1|k-1}^{a} & =\begin{bmatrix}P_{k-1|k-1} & 0_{(m_{x}-1)\times m_{n_{x}}}\\
				0_{m_{n_{x}}\times(m_{x}-1)} & C_{x,k}
			\end{bmatrix}
		\end{cases}$
		\item[{\footnotesize{}5:}] $\mathcal{X}_{k-1|k-1,0}^{a}=\hat{x}_{k-1|k-1}^{a}$
		\item[{\footnotesize{}6:}] \textbf{for} $j=\{1,2\ldots,(m_{a}-1)\}$
		\item[] $\qquad\delta\hat{x}_{k-1,j}^{a}:=\left(\sqrt{(m_{a}-1+\lambda)P_{k-1|k-1}^{a}}\right)_{j}$,
		see \eqref{eq:del_k-1_QNUKF}
		\item[] $\qquad\mathcal{X}_{k-1|k-1,j}^{a}=\hat{x}_{k-1|k-1}^{a}\oplus\delta\hat{x}_{j}^{a}$,
		see \eqref{eq:Sigma_QNUKF}
		\item[] $\qquad\mathcal{X}_{k-1|k-1,j+m_{a}}^{a}=\hat{x}_{k-1|k-1}^{a}\ominus\delta\hat{x}_{j}^{a}$,
		see \eqref{eq:Sigma_QNUKF}
		\item[] \textbf{end for}
		\item[{\footnotesize{}7:}] \textbf{for} $j=\{1,2\ldots,(m_{a}-1)\}$
		\item[] $\qquad\ensuremath{\qquad\mathcal{X}_{k|k-1,j}=\operatorname{f}(\mathcal{X}_{k-1|k-1,j}^{a},u_{k-1})}$
		\item[] \textbf{end for},\textbf{ }see \eqref{eq:propagate_sigma_points_augmented}
		\item[{\footnotesize{}8:}] $\hat{x}_{k|k-1}=\begin{bmatrix}\operatorname{QWA}(\{\mathcal{X}_{k|k-1,j,q}\},\{w_{j}^{m}\})\\
			{\displaystyle \sum_{j=0}^{2(m_{a}-1)}w_{j}^{m}\mathcal{X}_{k|k-1,j,-}}
		\end{bmatrix}$,\textbf{ }see \eqref{eq:state mean - QNUKF}
		\item[{\footnotesize{}9:}] $P_{k|k-1}=\sum_{j=0}^{2(m_{a}-1)}[w_{j}^{c}(\mathcal{X}_{k|k-1,j}\ominus\hat{x}_{k|k-1})(\mathcal{X}_{k|k-1,j}\ominus\hat{x}_{k|k-1})^{\top}]+C_{w,k}$,\textbf{
		}see \eqref{eq:state cov - QNUKF}
		\item[] /{*} Update step {*}/
		\item[{\footnotesize{}10:}] \textbf{for} $j=\{0,1,2\ldots,(m_{a}-1)\}$
		\item[] $\qquad\mathcal{Z}_{k|k-1,j}=h(\mathcal{X}_{k|k-1,j}^{a},f_{w})$
		\item[] \textbf{end for},\textbf{ }see \eqref{eq:propagate measurment}
		\item[{\footnotesize{}11:}] $\hat{z}_{k|k-1}=\sum_{j=0}^{2(m_{a}-1)}w_{j}^{m}\mathcal{Z}_{k|k-1,j}$,
		see \eqref{eq:zhat - QNUKF}
		\item[{\footnotesize{}12:}] $P_{z_{k},z_{k}}=\sum_{j=0}^{2(m_{a}-1)}w_{j}^{c}[\mathcal{Z}_{k|k-1,j}-\hat{z}_{k|k-1}][\mathcal{Z}_{k|k-1,j}-\hat{z}_{k|k-1}]^{\top}+C_{f}$,
		see \eqref{eq:P_zz - QUKF}
		\item[{\footnotesize{}13:}] $P_{x_{k},z_{k}}=\sum_{j=0}^{2(m_{a}-1)}w_{j}^{c}[\mathcal{X}_{k|k-1,j}\ominus\hat{x}_{k|k-1}][\mathcal{Z}_{k|k-1,j}-\hat{z}_{k|k-1}]^{\top}$,
		see \eqref{eq:P_xz - QUKF}
		\item[{\footnotesize{}14:}] $\begin{cases}
			K_{k} & =P_{x_{k},z_{k}}P_{z_{k},z_{k}}^{\top}\\
			\delta\hat{x}_{k|k-1} & =K(z_{k}-\hat{z}_{k|k-1})\\
			\hat{x}_{k|k} & =\hat{x}_{k|k-1}\oplus\delta\hat{x}_{k|k-1}\\
			P_{k|k} & =P_{k|k-1}-K_{k}P_{z_{k},z_{k}}K_{k}^{\top}
		\end{cases}$, see \eqref{eq:QNUKF_correction}
		\item[{\footnotesize{}15:}] $k=k+1$
	\end{enumerate}
	\textbf{end while}
\end{algorithm}

\section{Implementation and Results\label{sec:Results}}

To evaluate the effectiveness of the proposed QNUKF, the algorithm
is tested using the real-world EuRoC dataset of drone flight in 6
DoF \cite{Burri25012016}. For video of the experiment, visit the
following \href{https://youtu.be/CP3xiOcGrTc}{link}. This dataset
features an Asctec Firefly hex-rotor Micro Aerial Vehicle (MAV) flown
in a static indoor environment, from which IMU, stereo images, and
ground truth data were collected. The stereo images are two simultaneous
monochrome images obtained via an Aptina MT9V034 global shutter sensor
with a 20 Hertz frequency. Linear acceleration and angular velocity
of the MAV were measured by an ADIS16448 sensor with a 200 Hertz frequency.
Note that the data sampling frequency of the camera and IMU are different.
Consequently, the measurement data does not necessarily exist at each
IMU data sample point. To incorporate this, the system only performs
the update step when image data is retrieved, and sets $\hat{x}_{k|k}$
equal to the predicted estimated state vector $\hat{x}_{k|k-1}$ when
image data is unavailable.

\subsection{The Proposed QNUKF Output Performance}

\begin{figure}[h]
	\centering \includegraphics[width=1\columnwidth]{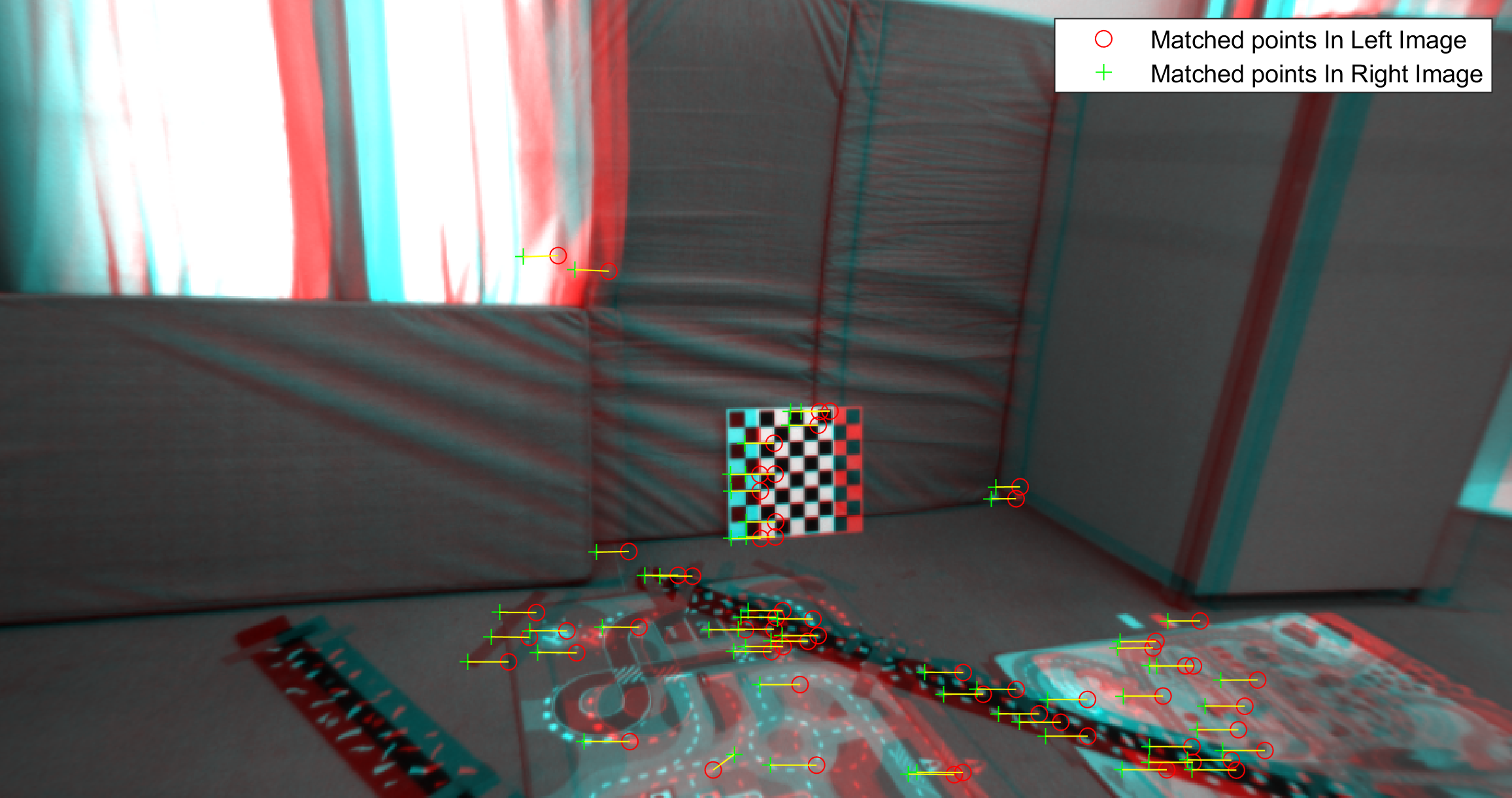} \caption{Matched feature points from the left and right views of the EuRoC
		dataset \cite{Burri25012016}.}
	\label{fig:matched}
\end{figure}

The feature point count $m_{f}$ is not constant at each measurement
step, to compensate for this, we set the measurement noise covariance
to:
\begin{equation}
	C_{f}=c_{f}^{2}I_{m_{f}}\label{eq:C_f}
\end{equation}
where $m_{f}$ is based on the number of feature points at each measurement
step and $c_{f}$ is a scaler tuned beforehand. To ensure the covariance
matrices remain symmetric and avoid numerical issues, for any $P\in\mathbb{R}^{m_{P}\times m_{P}}$,
the $\operatorname{Symmetrize}$ function is defined as $\operatorname{Symmetrize}:\mathbb{R}^{m_{P}\times m_{P}}\rightarrow\mathbb{R}^{m_{P}\times m_{P}}$:
\begin{equation}
	\operatorname{Symmetrize}(P):=\frac{P+P^{\top}}{2}\in\mathbb{R}^{m_{P}\times m_{P}}.\label{eq:Symmetrize}
\end{equation}
This function is applied to all single-variable covariance matrices
after they are calculated according to the QNUKF. The filter parameters
are configured as follows: $\lambda=3-(m_{a}-1)$, $\alpha=10^{-4}$,
and $\beta=2$. The IMU bias noise covariances are determined by the
standard deviations (std) of the bias white noise terms, set to $0.01\%$
of their average values to account for the slow random walk relative
to the IMU sampling rate. These values are configured as: $C_{ba,k}=10^{-8}\text{diag}(0.0022^{2},0.0208^{2},0.0758^{2})$
$(m/s^{2})^{2}$ and $C_{b\omega,k}=10^{-8}\text{diag}(0.0147^{2},0.1051^{2},0.0930^{2})$
$(rad/s)^{2}$. The std of the additive white noise terms are set
to $1\%$ of the measured vectors (i.e., linear acceleration and angular
velocity) such that the covariance matrices are $C_{\omega,k}=10^{-4}\text{diag}(0.1356^{2},0.0386^{2},0.0242^{2})$
$(m/s^{2})^{2}$ and $C_{a,k}=10^{-4}\text{diag}(9.2501^{2},0.0293^{2},-3.3677^{2})$
$(rad/s)^{2}$. The measurement noise std is calculated using the
measurement vectors from the V1\_01\_hard dataset \cite{Burri25012016},
recorded in the same room as the validation V1\_02\_medium dataset
\cite{Burri25012016}, but with different drone paths and complexity
speeds. Its value is found as $c_{f}=0.099538$ m. For each set of
stereo images obtained, feature points are identified using the Kanade-Lucas-Tomasi
(KLT) algorithm \cite{shi1994good}. An example of the feature match
results from this operation is depicted in Fig. \ref{fig:matched}.

Using triangulation \cite{hartley2003multiple}, the 2D matched points
are projected to the 3D space, representing the feature points in
$\{\mathcal{W}\}$. Using the $\mathbb{S}^{3}\times\mathbb{S}^{3}\rightarrow\mathbb{R}^{3}$
subtraction operator defined in \eqref{eq:q-q}, let us define the
orientation estimation error $r_{e,k}$ as:  
\begin{equation}
	r_{e,k}:=q_{k}\ominus\hat{q}_{k}\label{eq:r_e}
\end{equation}
where $r_{e,k}=\begin{bmatrix}r_{e1,k} & r_{e2,k} & r_{e3,k}\end{bmatrix}^{\top}\in\mathbb{R}^{3}$,
with each component $r_{ei,k}\in\mathbb{R}$ representing a dimension
of the error vector. Similarly, let us define the position and linear
velocity estimation errors at the $k$th time step, represented by
$p_{e,k}$ and $v_{e,k}$, respectively, as:  
\begin{align}
	p_{e,k} & :=p_{k}-\hat{p}_{k}\label{eq:p_e}\\
	v_{e,k} & :=v_{k}-\hat{v}_{k}\label{eq:v_e}
\end{align}
\begin{figure*}[!t]
	\centering \includegraphics[width=2\columnwidth]{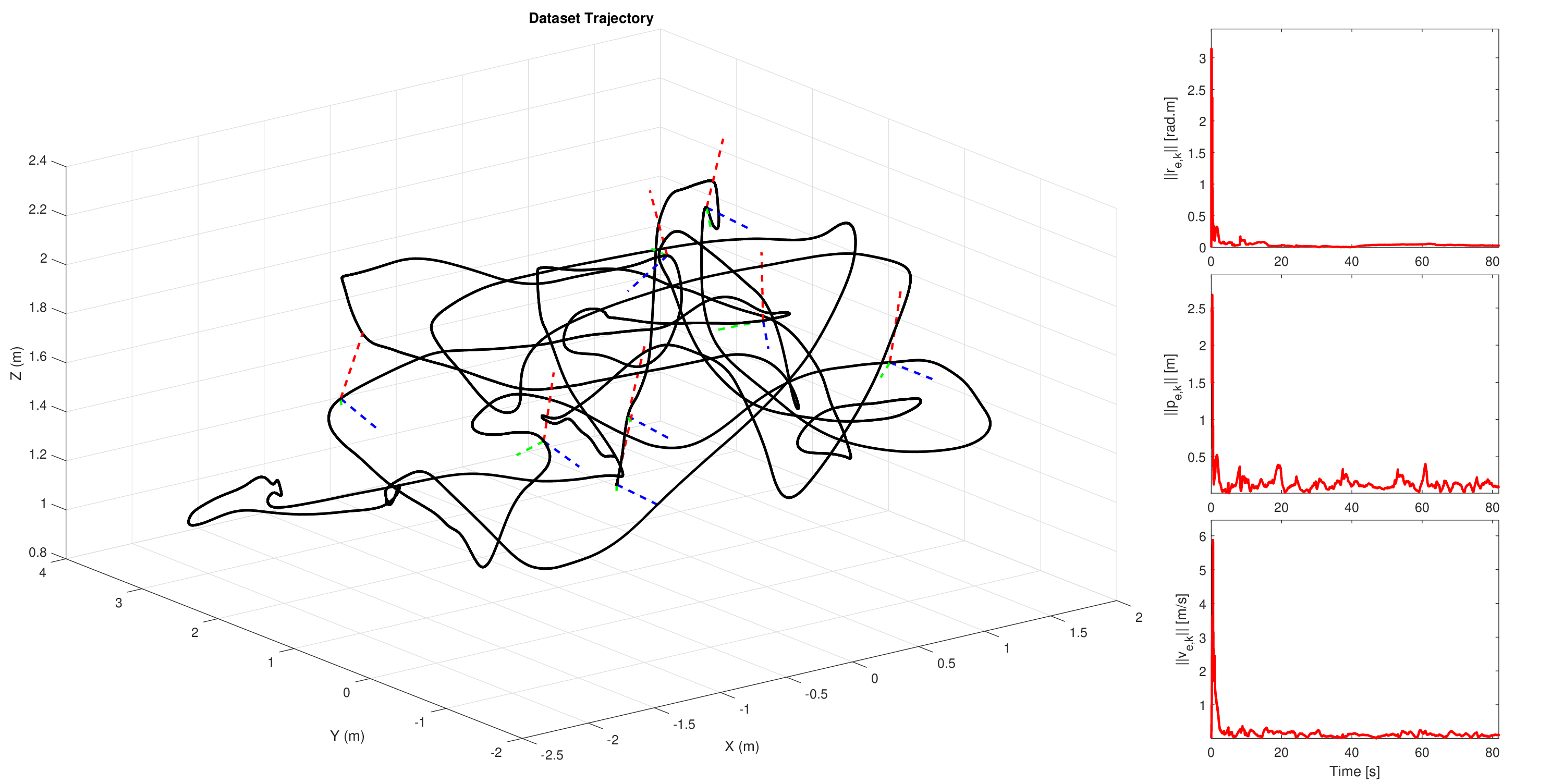}
	\caption{Validation results of the proposed QNUKF using the EuRoC V1\_02\_medium
		dataset \cite{Burri25012016}. On the left, the navigation trajectory
		of the drone moving in 3D space is shown with a solid black line,
		and the drone orientation with respect to $x$, $y$, and $z$ axes
		is represented by red, green, and blue dashed lines, respectively.
		On the right, the magnitudes of the orientation error vector $||r_{e,k}||$,
		position error vector $||p_{e,k}||$, and linear velocity error vector
		$||v_{e,k}||$ are plotted over time in solid red lines.}
	\label{fig:summary}
\end{figure*}

\begin{figure}[!h]
	\centering \includegraphics[width=1\columnwidth]{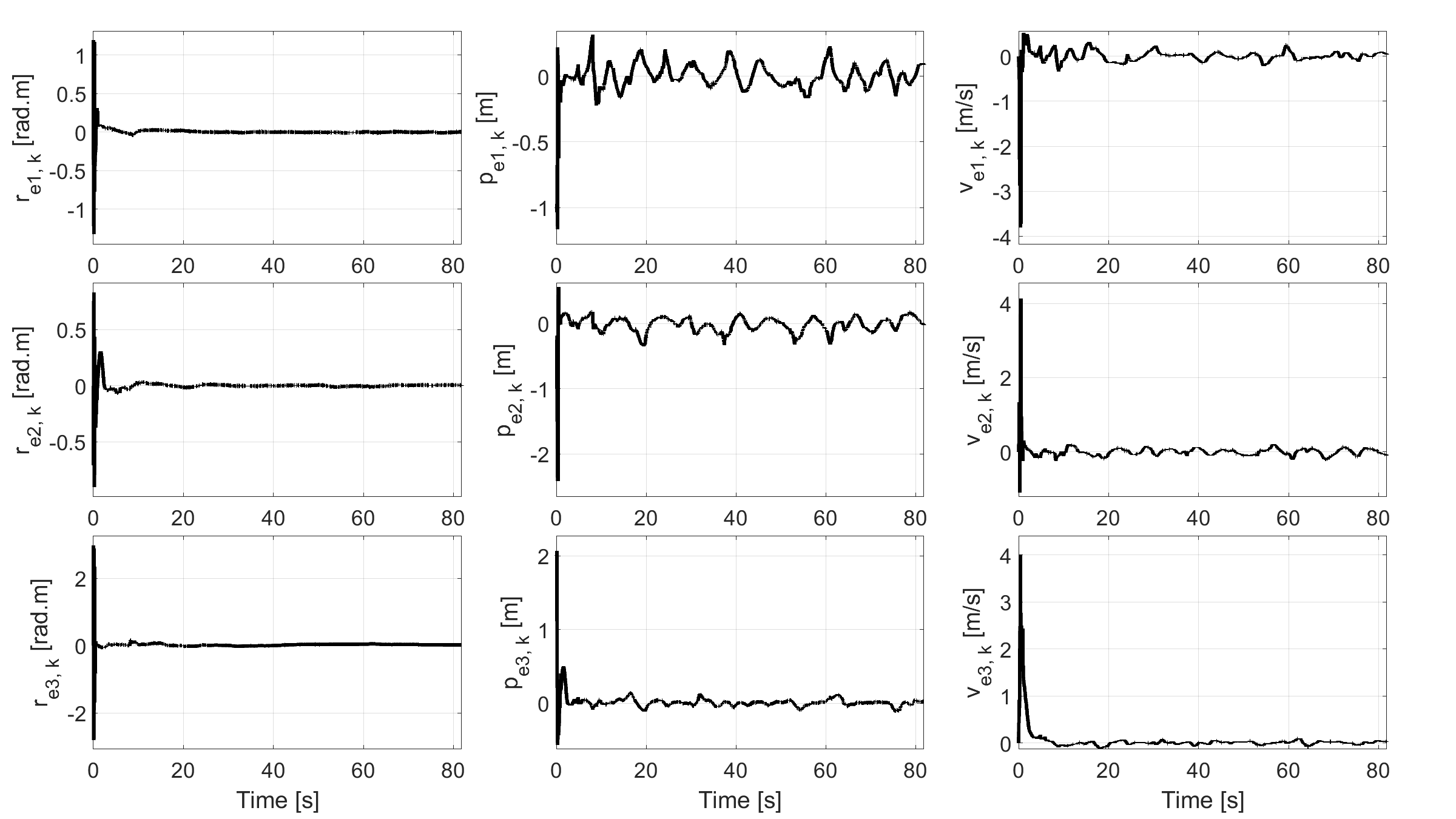}
	\caption{Estimation errors in rotation vector (left), position (middle), and
		linear velocity (right) components, from top to bottom of the proposed
		QNUKF.}
	\label{fig:errors}
\end{figure}
where $p_{e,k}=\begin{bmatrix}p_{e1,k} & p_{e2,k} & p_{e3,k}\end{bmatrix}^{\top}\in\mathbb{R}^{3}$
and $v_{e,k}=\begin{bmatrix}v_{e1,k} & v_{e2,k} & v_{e3,k}\end{bmatrix}^{\top}\in\mathbb{R}^{3}$.
The robustness and effectiveness of the proposed QNUKF have been
confirmed through experimental results on the EuRoC V1\_02\_medium
dataset \cite{Burri25012016} as depicted in Fig. \ref{fig:summary}
and with the initial estimates for the state vector and covariance
matrix being detailed as follows: $\hat{x}_{0:0}=[$$0.1619$, $0.7900$,
$-0.2053$, $0.5545$, $0.6153$, $2.0967$, $0.7711$, $0$, $0$,
$0$, $-0.0022$, $0.0208$, $0.0758$, $-0.0147$, $0.1051$, $0.0930$$]^{\top}$
and $P_{0|0}=\text{diag}(80I_{3},10I_{3},70I_{3},10I_{6})$.  On the
left side of Fig. \ref{fig:summary}, the trajectory and orientation
of the drone during the experiment are depicted. On the right side,
the magnitudes of the orientation, position, and linear velocity error
vectors are plotted over time. The results demonstrate that all errors
converge to near-zero values and recover from initial large errors
rapidly, indicating excellent performance. To examine the filter's
performance in detail, each component of the estimation error is plotted
against time. These plots are displayed in Fig. \ref{fig:errors},
showing excellent performance in tracking each component of orientation,
position, and linear velocity.

\subsection{Comparison with State-of-the-art Literature}

For video of the comparison, visit the following \href{https://youtu.be/BWraOI0LAXo}{link}.
To evaluate the efficacy of the proposed filter, we compared its performance
with the vanilla Extended Kalman Filter (EKF), the standard tool in
non-linear estimation problems as well as the Multi-State Constraint
Kalman Filter (MSCKF) \cite{sun2018robust}, which addresses the observability
issues of the EKF caused by linearization. Both models were implemented
to operate with the same state vectors (as defined in \eqref{state}),
state transition function (as defined in \eqref{state transition}),
and measurement function (as defined in \eqref{eq:measurement}).
Additionally, their noise covariance parameters, as well as initial
state and covariance matrices, were set to identical values in each
experiment for all the filters to ensure a fair comparison. In light
of \eqref{eq:r_e}, \eqref{eq:p_e}, and \eqref{eq:v_e}, let us define
the estimation error at time step $k$ denoted by $e_{k}$ as: 
\begin{equation}
	e_{k}=||r_{e,k}||+||p_{e,k}||+||v_{e,k}||\in\mathbb{R}\label{eq:error}
\end{equation}
Using \eqref{eq:error}, the Root Mean Square Error (RMSE) is calculated
as: 
\[
\text{RMSE}=\sqrt{\frac{1}{m_{k}}\sum_{k=1}^{m_{k}}e_{k}^{2}}
\]
where $m_{k}$ is the total number of time steps. The steady state
root mean square error (SSRMSE) is defined similarly to RMSE but calculated
over the last 20 seconds of the experiment. Table \ref{tab:all_comp}
compares the performance of the EKF, MSCKF, and QNUKF across three
different EuRoC \cite{Burri25012016} datasets using RMSE and SSRMSE
as metrics. Experiments 1, 2, and 3 correspond to the V1\_02\_medium,
V1\_03\_difficult, and V2\_01\_easy datasets, respectively, with the
best RMSE and SSRMSE values in each experiment. The proposed QNUKF
algorithm consistently outperformed the other filters across all experiments
and metrics. While the MSCKF showed better performance than the EKF,
it lacked consistency, with RMSE values ranging from $0.866786$ to
$1.454449$. In contrast, the QNUKF maintained a more stable performance,
with RMSE values tightly clustered between $0.265037$ and $0.331952$.

\begin{table}
	\centering \caption{Performance Comparison of QNUKF (proposed) vs EKF and MSCKF (literature)
		on EuRoC Datasets.}
	\label{tab:all_comp} 
	\global\long\def\arraystretch{1.2}%
	\setlength{\arrayrulewidth}{1pt} 
	\begin{tabular}{llccc}
		\hline 
		\textbf{\setlength{\arrayrulewidth}{0.4pt} 
		}Filter & Metric & Experiment 1 & Experiment 2 & Experiment 3\tabularnewline
		\hline 
		\multirow{2}{*}{EKF} & \multicolumn{1}{l}{RMSE} & $0.952955$ & $0.831777$ & $1.411102$\tabularnewline
		& \multicolumn{1}{l}{SSRMSE} & $0.123161$ & $0.160883$ & $0.263435$\tabularnewline
		\hline 
		\multirow{2}{*}{MSCKF} & \multicolumn{1}{l}{RMSE} & $0.866786$ & $1.175951$ & $1.454449$\tabularnewline
		& \multicolumn{1}{l}{SSRMSE} & $0.074154$ & $0.062882$ & $0.075329$\tabularnewline
		\hline 
		\multirow{2}{*}{QNUKF} & \multicolumn{1}{l}{RMSE} & $0.331952$ & $0.275067$ & $0.265037$\tabularnewline
		& \multicolumn{1}{l}{SSRMSE} & $0.059464$ & $0.051633$ & $0.060865$\tabularnewline
		\hline 
	\end{tabular}
\end{table}

\begin{figure}[!h]
	\centering \includegraphics[width=1\columnwidth]{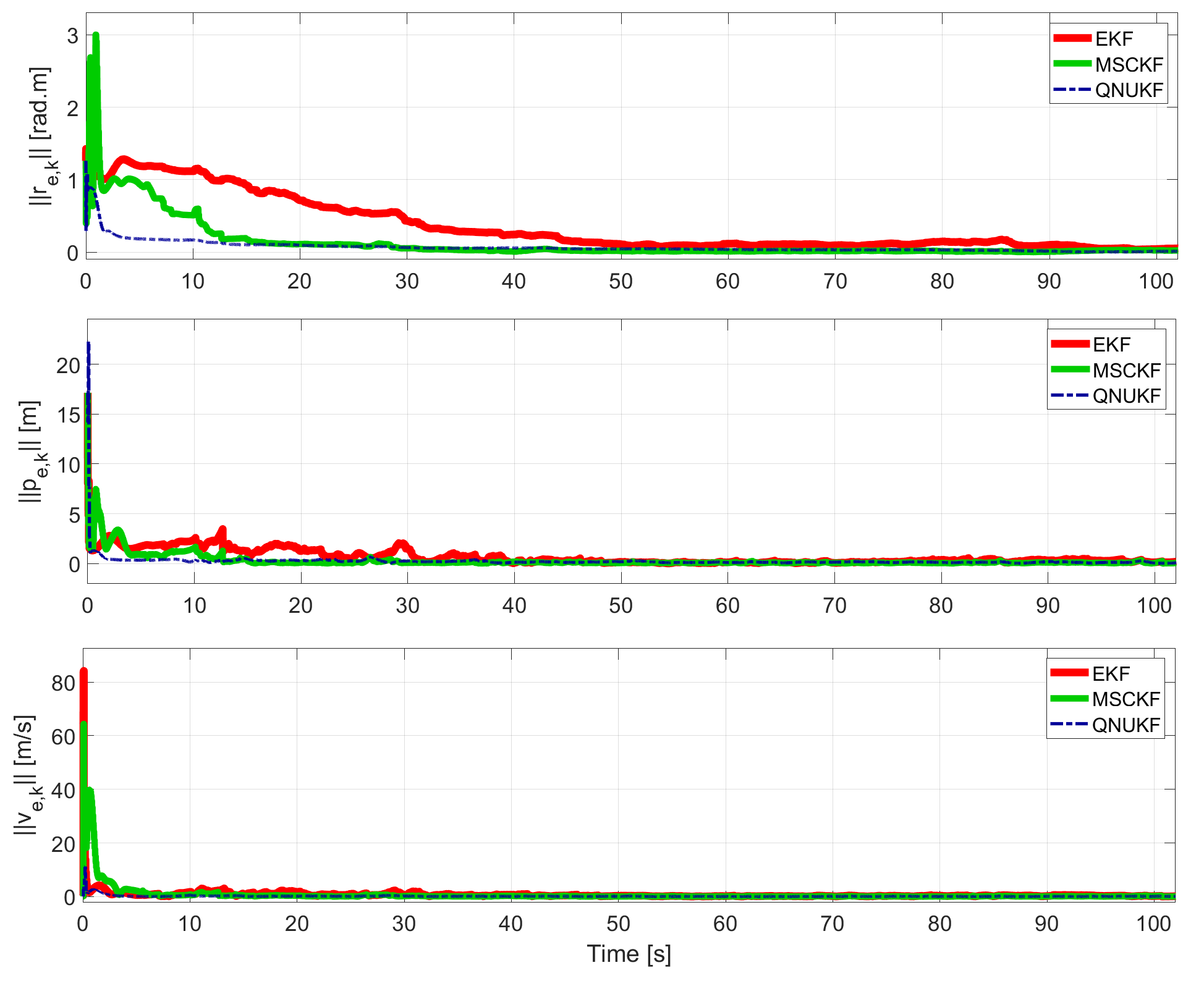}
	\caption{Estimation errors in orientation (top), position (middle), and linear
		velocity (bottom) for the EKF (red thick solid line), MSCKF (green
		solid line), and proposed QNUKF (blue dashed line).}
	\label{fig:comp_errors}
\end{figure}

Let us examine the orientation, position, and linear velocity estimation
error trajectories for all the filters in Experiment 2 as a case study.
As depicted in Fig. \ref{fig:comp_errors}, the QNUKF outperformed
the other filters across all estimation errors. The EKF exhibited
a slow response in correcting the initial error, while the MSCKF corrected
course more quickly but initially overshot compared to the EKF. In
contrast, the QNUKF responded rapidly and maintained smaller steady-state
errors when compared with EKF and MSCKF.


\section{Conclusion \label{sec:Conclusion}}

In this paper, we investigated the orientation, position, and linear
velocity estimation problem for a rigid-body navigating in three-dimensional
space with six degrees-of-freedom. A robust Unscented Kalman Filter
(UKF) is developed, incorporating quaternion space, ensuring computational
efficiency at low sampling rates, handling kinematic nonlinearities
effectively, and avoiding orientation singularities, such as those
common in Euler angle-based filters. The proposed Quaternion-based
Navigation Unscented Kalman Filter (QNUKF) relies on onboard Visual-Inertial
Navigation (VIN) sensor fusion of data obtained by a stereo camera
(feature observations and measurements) and a 6-axis Inertial Measurement
Unit (IMU) (angular velocity and linear acceleration). The performance
of the proposed algorithm was validated using real-world dataset of
a drone flight travelling in GPS-denied regions where stereo camera
images and IMU data were collected at low sampling rate. The results
demonstrated that the algorithm consistently achieved excellent performance,
with orientation, position, and linear velocity estimation errors
quickly converging to near-zero values, despite significant initial
errors and the use of low-cost sensors with high uncertainty. The
proposed algorithm consistently outperformed both baseline and state-of-the-art
filters, as demonstrated by the three experiments conducted in two
different rooms.

As the next step, we will investigate the effectiveness of adaptive
noise covariance tuning on QNUKF to mitigate this issue. Additionally,
reworking QNUKF to address simultaneous localization and mapping (SLAM)
for UAVs is a consideration for future work, as it represents a natural
progression from navigation.

\balance
\bibliographystyle{IEEEtran}
\bibliography{bib_QUKF_ACC}
		
	\end{document}